\documentclass{article}

\usepackage{microtype}
\usepackage{graphicx}
\usepackage{booktabs} %

\usepackage{hyperref}

\usepackage[accepted]{icml2024}

\usepackage{amsmath}
\usepackage{amssymb}
\usepackage{mathtools}
\usepackage{amsthm}
\usepackage{subcaption}
\usepackage{textcomp}
\usepackage{tablefootnote}

\usepackage[capitalize,noabbrev]{cleveref}
\usepackage{subfiles} %

\theoremstyle{plain}
\newtheorem{theorem}{Theorem}[section]

\theoremstyle{definition}

\theoremstyle{remark}

\DeclareMathOperator*{\argmin}{arg\,min}
\newcommand{\Dtr}{D_{\text{tr}}}
\newcommand{\Dtropt}{D_{\text{tr}}^*}

\newcommand{\Dinfi}{D_{\text{inf}}}
\newcommand{\Dinp}{D_{\text{inp}}}
\newcommand{\Dout}{D_{\text{out}}}
\DeclareMathOperator*{\Tf}{T_{\text{FLOPs}}}
\DeclareMathOperator*{\If}{I_{\text{FLOPs}}}

\DeclareMathOperator*{\Utr}{MFU_\text{tr}}
\DeclareMathOperator*{\Uinp}{MFU_\text{inp}}
\DeclareMathOperator*{\Uout}{MFU_\text{out}}
\newcommand{\Ctr}{C_\text{tr}}
\newcommand{\Cinf}{C_\text{inf}}

\renewcommand{\textapprox}{\raisebox{0.5ex}{\texttildelow}}

\usepackage[textsize=tiny]{todonotes}

\icmltitlerunning{Accounting for Inference in Scaling Laws}

\begin{document}

\twocolumn[
\icmltitle{Beyond Chinchilla-Optimal:\\
Accounting for Inference in Language Model Scaling Laws}

\icmlsetsymbol{equal}{*}

\begin{icmlauthorlist}
\icmlauthor{Nikhil Sardana}{comp}
\icmlauthor{Jacob Portes}{comp}
\icmlauthor{Sasha Doubov}{comp}
\icmlauthor{Jonathan Frankle}{comp}
\end{icmlauthorlist}

\icmlaffiliation{comp}{Databricks MosaicML, United States of America}

\icmlcorrespondingauthor{Nikhil Sardana}{nikhil@mosaicml.com}

\icmlkeywords{Scaling Laws, LLMs, Language Models, Machine Learning, ICML}

\vskip 0.3in
]

\printAffiliationsAndNotice{}  %

\begin{abstract}
 Large language model (LLM) scaling laws are empirical formulas that estimate changes in model quality as a result of increasing parameter count and training data. However, these formulas, including the popular Deepmind Chinchilla scaling laws, neglect to include the cost of inference. We modify the Chinchilla scaling laws to calculate the optimal LLM parameter count and pre-training data size to train and deploy a model of a given quality and inference demand. We conduct our analysis both in terms of a compute budget and real-world costs and find that LLM researchers expecting reasonably large inference demand (\textapprox 1B requests) should train models smaller and longer than Chinchilla-optimal. Furthermore, we train 47 models of varying sizes and parameter counts to validate our formula and find that model quality continues to improve as we scale tokens per parameter to extreme ranges (up to 10,000). Finally, we ablate the procedure used to fit the Chinchilla scaling law coefficients and find that developing scaling laws only from data collected at typical token/parameter ratios overestimates the impact of additional tokens at these extreme ranges.
\end{abstract}

\section{Introduction}
\label{sec:intro}

Large language models (LLMs) have substantial training and inference compute and energy costs \cite{openaiinterview, palminference}. Training computation costs are primarily determined by the size of the model and the amount of data seen during training \cite{chinchilla}. For state-of-the-art models with tens of billions of parameters trained on trillions of tokens, training costs can easily exceed millions of dollars. 
Similarly, inference costs depend on the size of the model and the volume of user queries over the lifetime of the model. This volume can be significant; demand for popular models can exceed billions of tokens per day \cite{openaidemand, characterai}.

Accounting for both \textit{training and inference}, how does one minimize the cost required to produce and serve a high quality model?

Recent studies have proposed scaling laws, empirical formulas that estimate how changes in model and training data size impact model quality \citep{scalinglaws, chinchilla}. \citet{chinchilla} is perhaps the most influential of these works, finding that to scale language models most efficiently, parameters and tokens should grow approximately linearly. The authors applied this scaling law to train a 70B parameter model (dubbed \textit{Chinchilla}) that outperformed much larger and more expensive models such as GPT-3. As a result, many subsequent LLMs have been trained following the Chinchilla scaling laws \citep{cerebrasgpt, dataconstrained}.

However, the Chinchilla scaling laws only account for the computational costs of training. By contrast, the Llama 2 family of models were trained on 2 trillion tokens and the Llama 3 family of models were trained on 15 trillion tokens, which is far more data than the Chinchilla scaling laws would deem ``optimal'' \citep{llama, llama2, llama3}. Since inference costs are lower for smaller models, the extra training compute required to train a Llama-style model over a Chinchilla-style model of equivalent quality pays off after enough inference requests.

\begin{figure*}
        \begin{subfigure}[b]{0.33\textwidth}
                \centering
            \includegraphics[width=\linewidth]{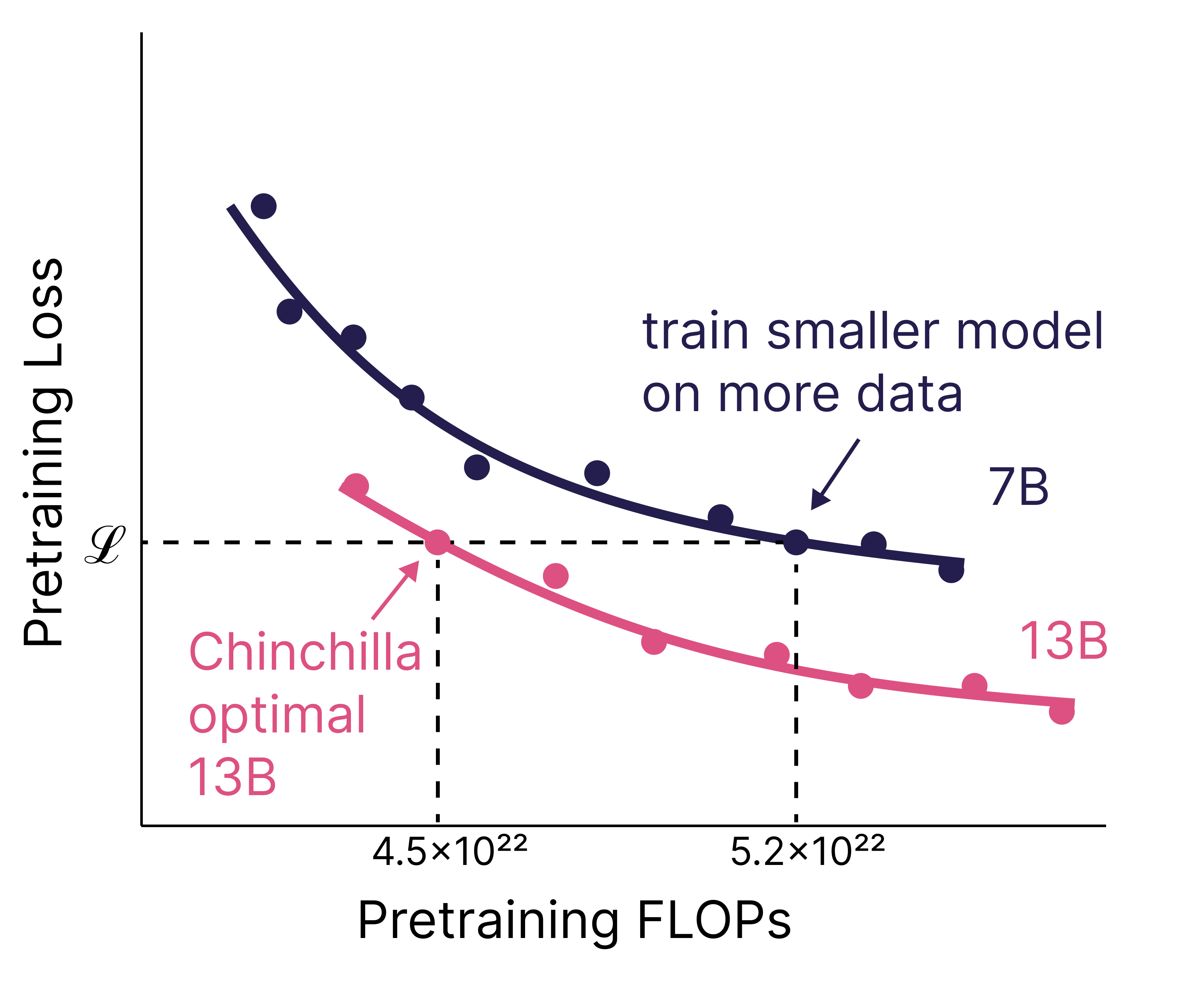}
                \caption{}
        \end{subfigure}%
        \begin{subfigure}[b]{0.33\textwidth}
                \centering
                \includegraphics[width=\linewidth]{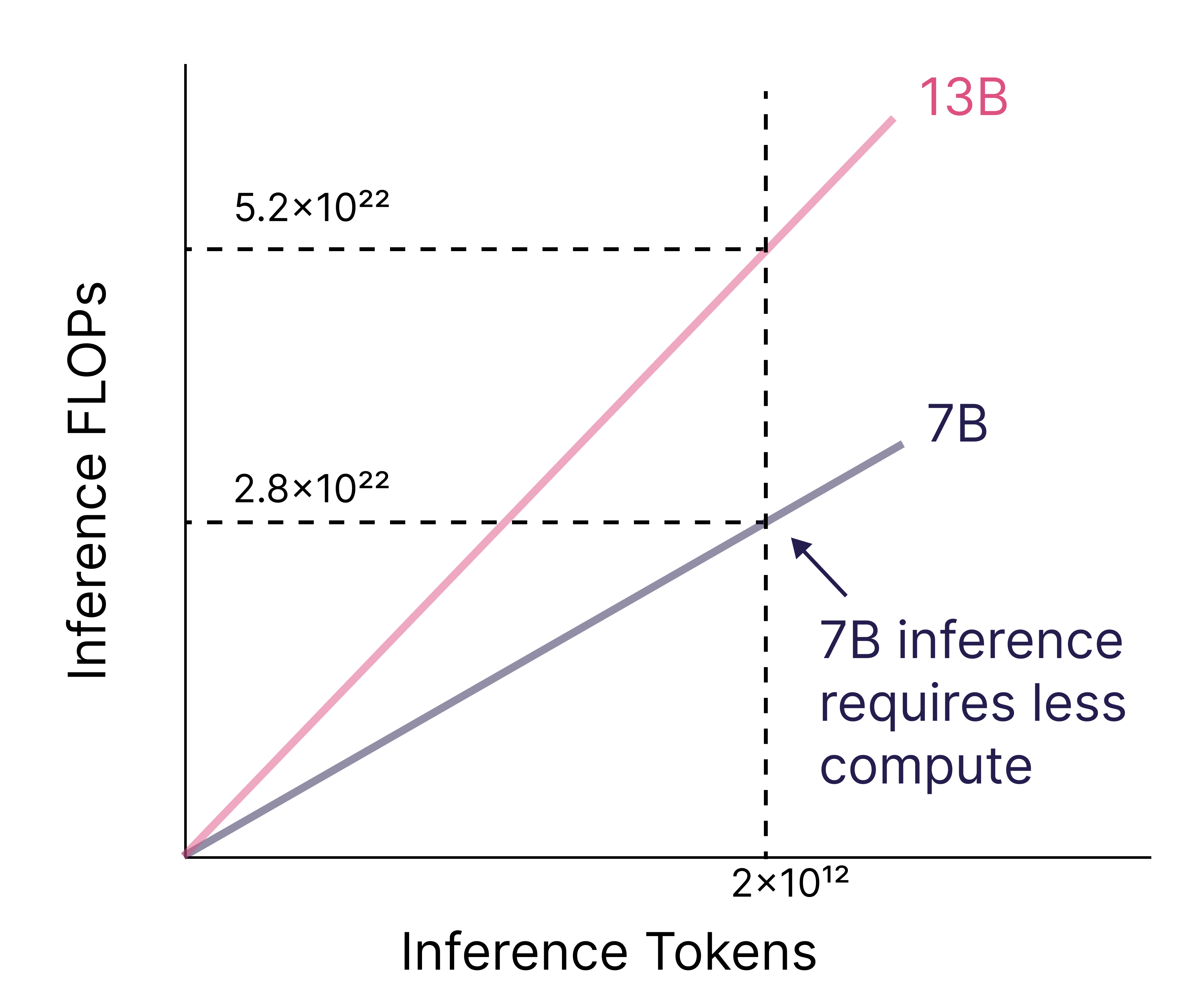}
                \caption{}
        \end{subfigure}%
        \begin{subfigure}[b]{0.33\textwidth}
                \centering
                \includegraphics[width=\linewidth]{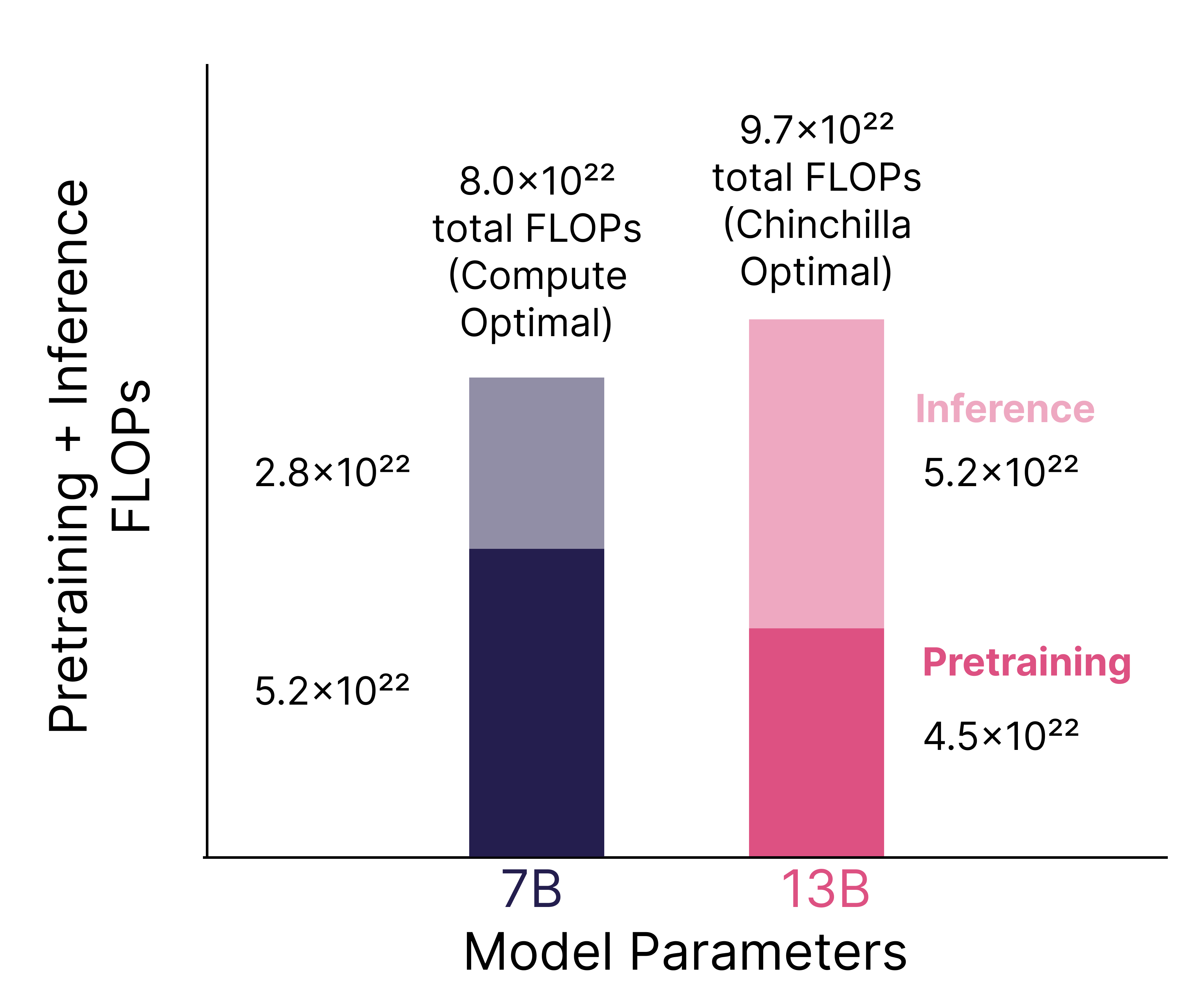}
                \caption{}
        \end{subfigure}%
        \caption{Schematic of compute savings achieved via our approach. An LLM developer seeking to train a 13B model who expects 2 trillion tokens of inference demand during the model's lifetime can reduce their total compute by $1.7\times10^{22}$ FLOPs ($17\%$) by instead training a 7B model on more data (a). The extra compute required to train the 7B model beyond its Chinchilla-optimal point to match the 13B's quality is made up for during inference (b), (c). Our method quantifies this training-inference trade-off, producing models that are optimal over their total lifetime.}
    \label{fig:hero}
\end{figure*}

Prior work has discussed the training-inference compute trade-off \citep{llama, llama2, stablelm, blogpost, tradeoff}. \citet{llama} cites the lower inference cost of smaller models as inspiration for the LLaMA series. \citet{blogpost} calculates the compute overhead of training longer than Chinchilla, but does not discuss quantify compute savings from inference. \citet{tradeoff} discuss this trade-off in more detail, but show the shift in scaling laws for only a single particular number of inferences.

Other related work includes \citet{dataconstrained}, which adapts the Chinchilla scaling laws for the data-constrained regime, where we have to repeat training tokens. Our problem setting is the opposite: We assume we are \textit{data rich} but \textit{compute constrained}, and seek to minimize computation costs assuming we have enough data to train high-quality models. 

In this paper, we modify the Chinchilla scaling laws to account for inference costs by calculating the optimal parameter and training token counts---both in terms of compute (Sec. \ref{sec:compopt}) and dollar costs (Sec. \ref{sec:costopt})---to train and deploy a model of any given quality and inference demand. Our principled derivation estimates that LLM practitioners expecting significant demand (\textapprox$10^9$ inference requests) should train models substantially smaller and longer than Chinchilla-optimal. Figure \ref{fig:hero} illustrates the benefits of our \textit{compute-optimal} method for a realistic scenario.

In inference-heavy regimes, our modification predicts that it is optimal to train a model far smaller and longer than Chinchilla predicts, up to thousands of tokens per parameter. Does this hold true in practice? Do transformer models see continued improvements at such extreme cases? Or is there a point beyond which models saturate, and additional tokens provide no further improvement, as \citet{blogpost} suggests? To uncover the behavior of transformer models in these cases, we train 47 models ranging from 150M to 6B parameters, on various data budgets from 10 to 10,000 tokens per parameter. We find that model quality continues to improve as we scale token ratios. We do not find evidence of a ``saturation point," beyond which models do not improve, even with additional data.

Lastly, we ablate the parametric curve fitting procedure from the Chinchilla scaling laws. The Chinchilla scaling laws use empirical data from over 400 training runs to determine coefficients that estimate precisely how additional model parameters and training data impact loss. \citet{chinchilla} collected data only from standard token ratio training runs ($\leq$\textapprox 100 tokens/parameters). Our ablation indicates that when fitting Chinchilla coefficients using only typical token ratio runs, this formula \textit{overestimates} the impact of additional training data as we move to the extreme ratio regime.

\section{Computational Optimality}
\label{sec:compopt}

\begin{figure*}
        \begin{subfigure}[b]{0.33\textwidth}
                \centering
            \includegraphics[width=\linewidth]{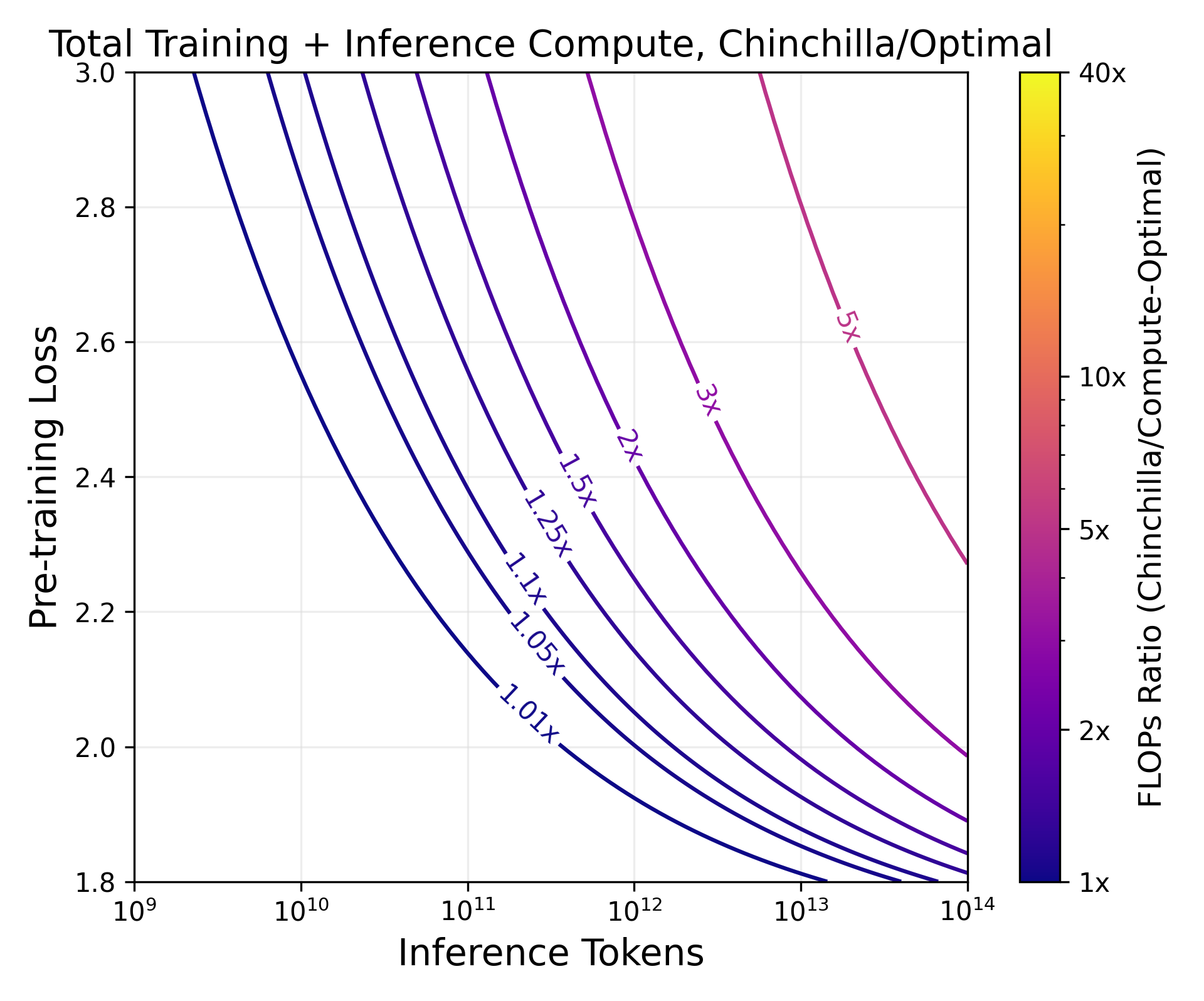}
                \caption{}
                \label{fig:flops}
        \end{subfigure}%
        \begin{subfigure}[b]{0.33\textwidth}
                \centering
                \includegraphics[width=\linewidth]{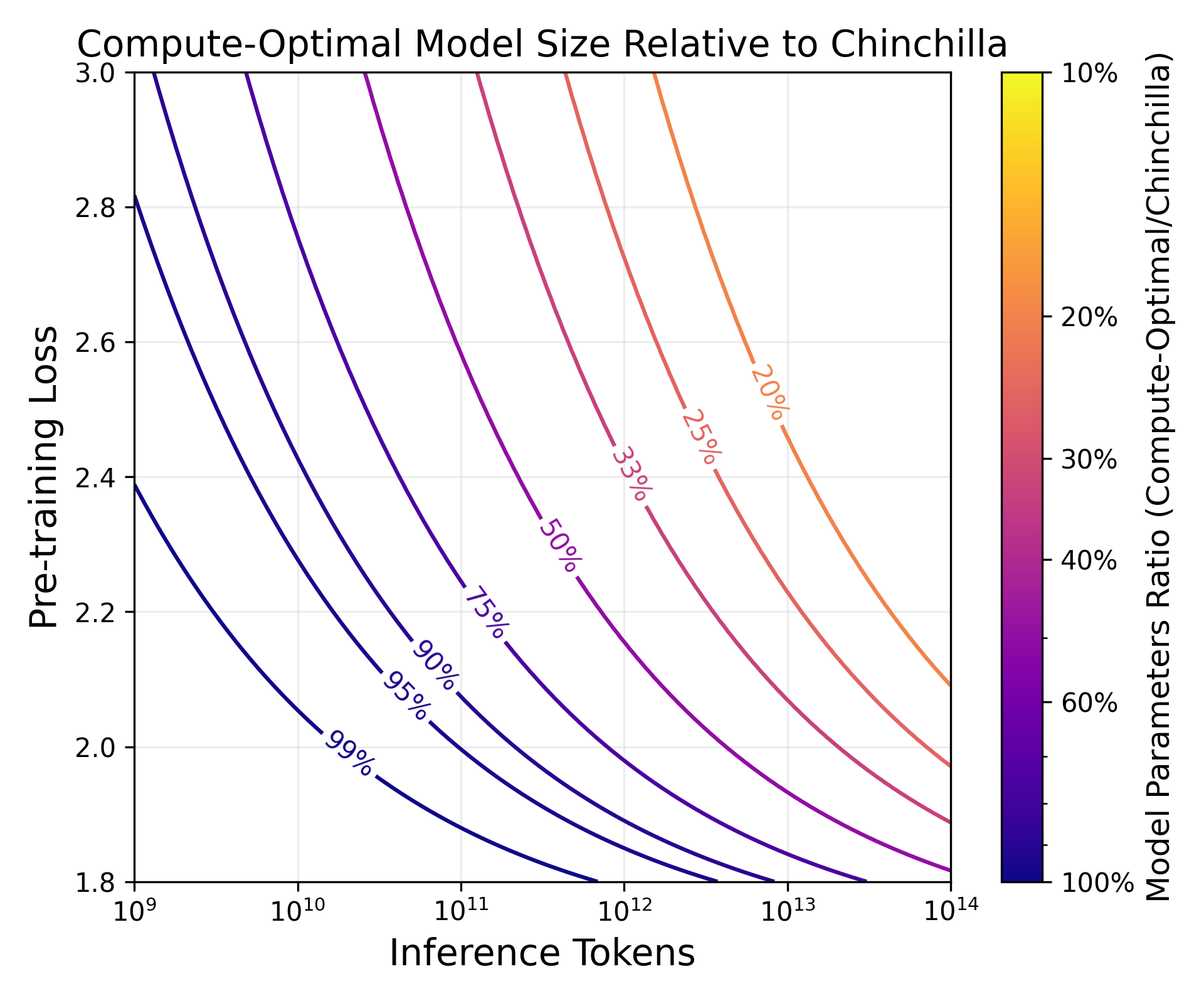}
                \caption{}
                \label{fig:params}
        \end{subfigure}%
        \begin{subfigure}[b]{0.33\textwidth}
                \centering
                \includegraphics[width=\linewidth]{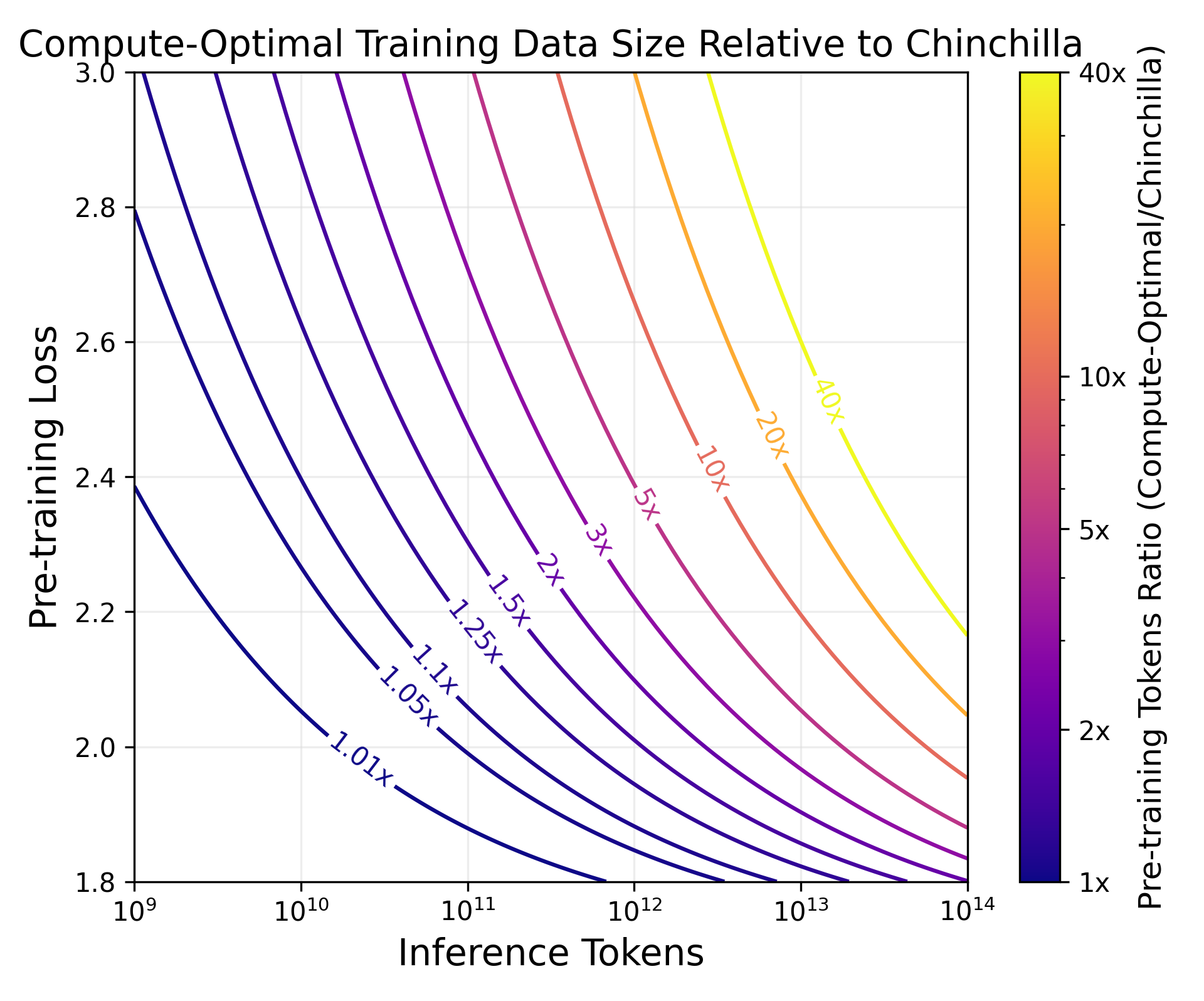}
                \caption{}
                \label{fig:tokens}
        \end{subfigure}%
        \caption{Ratios of (a) total FLOPs, (b) model parameters, and (c) pre-training tokens, for optimal models estimated via our method vs. Chinchilla-style models.
        For each point $(x, y)$ in the figures, we compute the Chinchilla model parameter count and training data required to reach the  loss $y$, and the number of combined FLOPs required to train and run inference for $x$ tokens using the Chinchilla model. Then, we compute the same values (total FLOPs, parameter count, training data size) for the compute-optimal models returned by our method, and plot the ratios.}
        \label{fig:ratios}
\end{figure*}

We seek to minimize the computational costs of a model of a given quality and inference demand. We closely follow the methodology in \citet{chinchilla} (henceforth referred to as ``the Chinchilla paper''), using pre-training cross-entropy loss as a proxy for quality, and floating-point operations (FLOPs) as our unit of computational cost. 

We model our pre-training loss $L(N, \Dtr)$ in terms of the number of parameters, $N$, and pre-training tokens, $\Dtr$, according to the Chinchilla paper's third scaling law:
\begin{align}\label{eq:loss}
 L(N, \Dtr) \triangleq E + \frac{A}{N^\alpha} + \frac{B}{D_{\text{tr}}^\beta}
\end{align}
The Chinchilla paper derived the parametric loss function in Eq. \ref{eq:loss} and fit values for $A, B, E, \alpha$, and $\beta$ from the authors' empirical training results.
The best-fit values for these coefficients depend on the exact dataset and model architecture; however, the Chinchilla paper found largely consistent results across the MassiveText, Github \citep{gopher}, and C4 \citep{c4} datasets, and subsequent work has replicated these scaling laws on other internet corpora and transformer variants \cite{cerebrasgpt,besiroglu2024chinchilla,gadre2024language}. We use the coefficient values from the Chinchilla paper in our analysis here, and explore fitting these coefficients on different datasets and data ratios more detail in Section \ref{sec:parametric}.

Additionally, we assume that conditioned on pre-training loss, inference demand is independent of model size and token count. In other words, models of equivalent quality but different parameter counts will see the same requests.\footnote{In practice, smaller models of equivalent quality may have greater demand since they can have lower inference latency.}

Let $\Tf(N, D)$ and $\If(N, D)$ be the number of FLOPs required to train and run inference, respectively, on a model with $N$ parameters for $D$ tokens. Denote the number of tokens (input + output) of a single inference request $i$ as $D_{\text{inf}}^{(i)}$. Let $\Dinfi = \sum_{i} D_{\text{inf}}^{(i)}$ be the sum of all tokens over all inference requests.

Formally, we are interested in minimizing the sum of our training and inference FLOPs under the constraint $L(N, \Dtr) = \ell$:
\begin{align}
\label{eq:1}
\begin{split}
N^*(\ell, \Dinfi), D_{\text{tr}}^{*}(\ell,\Dinfi) = &\argmin_{N, \Dtr \mid L(N, \Dtr) = \ell} \Tf(N, \Dtr)\\
&+ \sum_{i}^{} \If(N, \Dinfi^{(i)}).
\end{split}
\end{align}
$N^*$ and $D_{\text{tr}}^{*}$ are functions that describe the optimal parameters and pre-training tokens, respectively, that minimize total training and inference compute. The pre-training loss constraint ensures that we minimize compute for a given quality.

We use the standard approximation of FLOPs for transformer models with $N$ parameters: $6N$ per training token and $2N$ per inference token \citep{scalinglaws}. Thus, our objective simplifies to:
\begin{align}
\label{eq:2}
\begin{split}
N^*(\ell, \Dinfi), \Dtropt(\ell, \Dinfi) &=\\
&\argmin_{N, \Dtr \mid L(N, \Dtr) = \ell} 6N\Dtr + 2N\Dinfi.
\end{split}
\end{align}
We note that this is the ``converse'' of the Chinchilla optimization problem. In the Chinchilla paper, the authors assumed a \textit{fixed} compute budget and found $N^*$ and $\Dtropt$ that \textit{minimized} pre-training loss. Our objective is to \textit{fix} pre-training loss and find $N^*$ and $\Dtropt$ that \textit{minimize} compute costs.

Crucially, our total computational cost depends on the inference demand over the lifetime of the model, but our model's parameter count and data size are determined prior to training. Thus, our analysis is predicated on the assumption that LLM practitioners can estimate their inference demand prior to training.

Without inference ($\Dinfi = 0$), the optimization problem in Eq. \ref{eq:2} can be solved analytically. Unfortunately, accounting for inference ($\Dinfi > 0$), determining $N^*$ and $\Dtropt$ analytically as functions of $\ell$ and $\Dinfi$ is intractable (we defer our proof to Appendix \ref{sec:proof1}). Instead, we computationally solve for $N^*$ and $\Dtropt$ across a range of values of $\ell$ and $\Dinfi$ using the Newton root-finding method. In practice, this method converges for relevant inputs and we are able to determine optimal parameter/token counts.

\begin{figure*}[t!]
        \begin{subfigure}[b]{0.33\textwidth}
                \centering
                \includegraphics[width=1.00\linewidth]{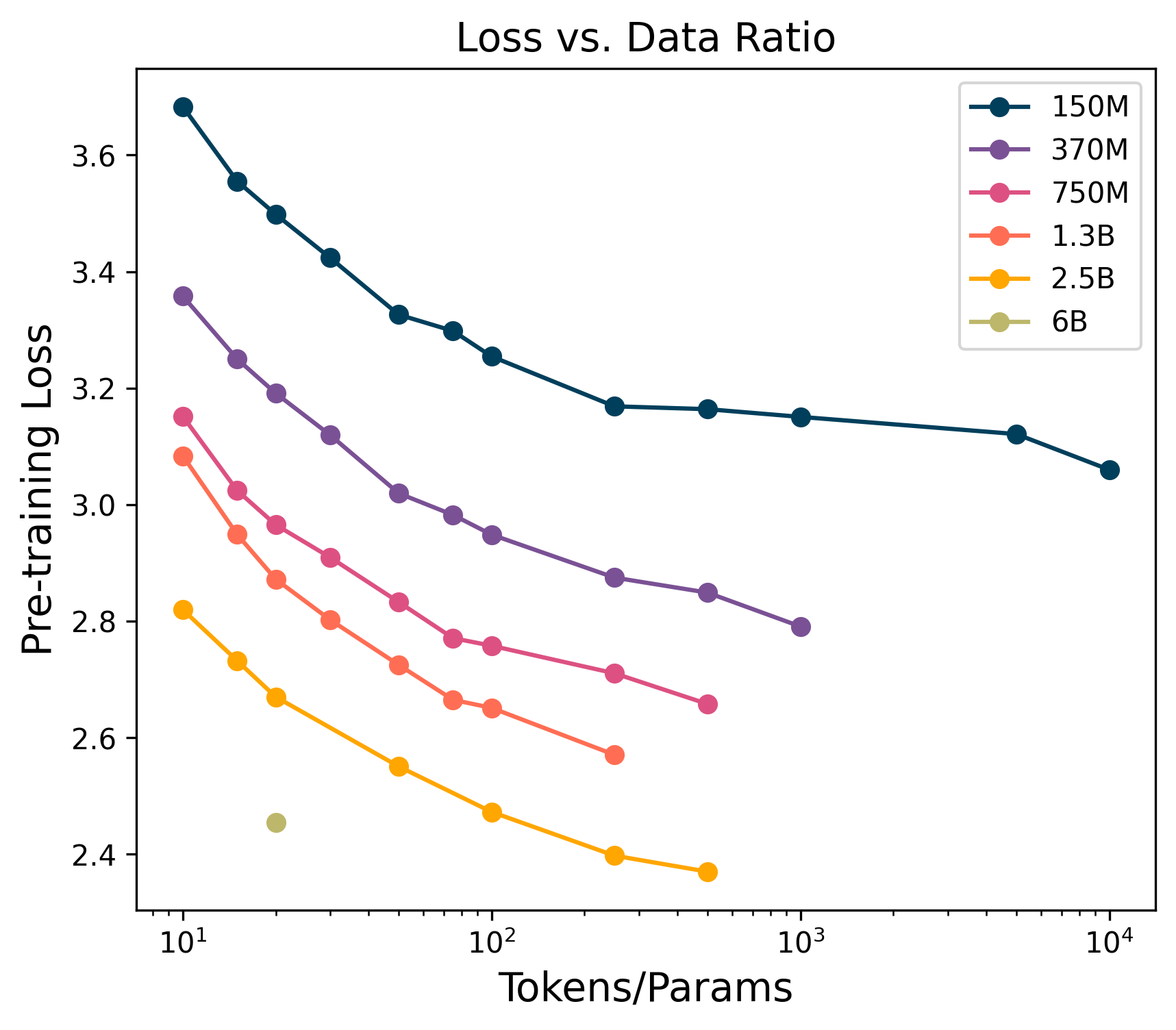}
                \caption{}
                \label{fig:loss}
        \end{subfigure}%
        \begin{subfigure}[b]{0.33\textwidth}
                \centering
                \includegraphics[width=1.00\linewidth]{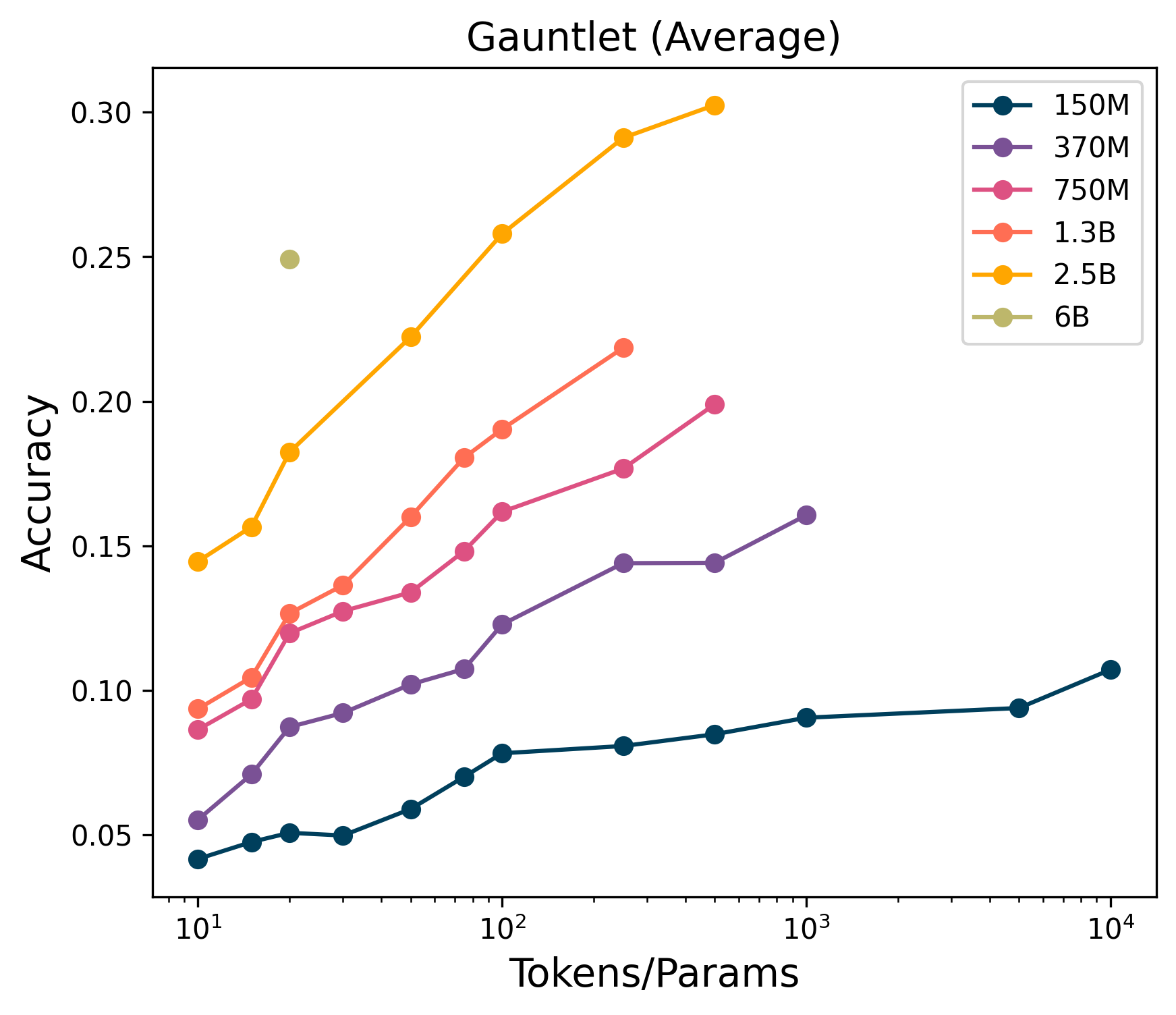}
                \caption{}
                \label{fig:gauntlet_tpr}
        \end{subfigure}%
        \begin{subfigure}[b]{0.33\textwidth}
                \centering
                \includegraphics[width=1.00\linewidth]{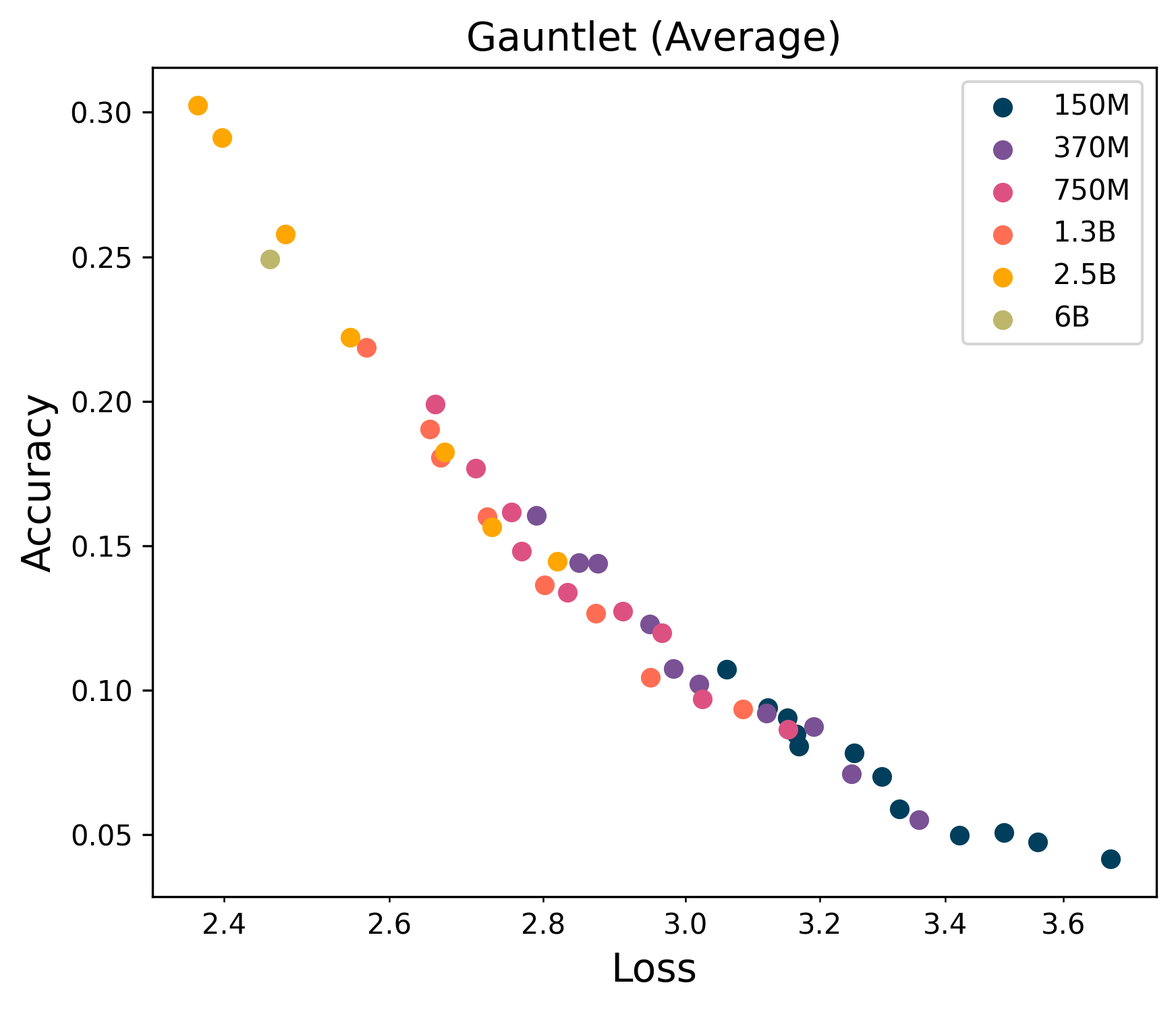}
                \caption{}
                \label{fig:gauntlet_loss}
        \end{subfigure}%
        \caption{For each model (150M, 370M, 750M, 1.3B, 2.5B, 6B) in our experimental sweep, we plot (a) Loss vs. Tokens per parameter, (b) Gauntlet Average (an aggregation of all our metrics described in Sec. \ref{sec:experiments}) vs. Tokens per parameter, and (c) Loss vs. Gauntlet Average. Category-specific Gauntlet results are available in Sec. \ref{sec:furtherresults}.}
        \label{fig:experimentresults}
\end{figure*}

In Figure \ref{fig:ratios}, we show how our inference-adjusted model's FLOP counts, parameters, and pre-training tokens compare to Chinchilla-style models across a range of loss values and inference demands. When inference usage is significantly less than the number of pre-training tokens, Chinchilla models are essentially compute-optimal. However, as demand increases, inference costs becomes a significant factor. For a 7B-Chinchilla-quality model with an inference demand of $10^{11}$ tokens, our formula suggests the compute-optimal method is to train a 6B parameter model on 1.18$\times$ the original (Chinchilla-prescribed) amount of data. For higher quality models (i.e. models that are larger and/or trained for longer), the volume of inference demand required to shift the scaling law increases: An LLM developer that expects a 30B-Chinchilla-quality model will see $10^{13}$ tokens during inference can reduce their total FLOPs by 28\% by training a 13.6B model on 2.84$\times$ the data. We provide additional results in Sec. \ref{sec:compoptappendix}
in the Appendix.

\section{Experiments}
\label{sec:experiments}
In high-demand inference scenarios, our analysis in Section \ref{sec:compopt} suggests that we should train models significantly smaller and on much more data than Chinchilla-style, resulting in hundreds or even thousands of tokens per parameter. The Chinchilla scaling laws have no upper bound---their form infinitely predicts that as tokens per parameter increase, model quality increases. However, the Chinchilla authors do not validate the scaling law at the outer ranges, conducting experiments only at typical ($<$\textapprox100 tokens/parameter) ratios. \citet{blogpost} postulates that there is a critical model size (\textapprox 30\%), below which, it is not possible to train on \textit{any} number of tokens and match a Chinchilla-style model.

To characterize the behavior of transformers at extreme data sizes, we train 47 models with the MPT architecture \cite{mpt} of varying size and token ratios. Our models range from 150M to 6B parameters, and our data budgets from 10 to 10,000 tokens per parameter. Due to resource constraints, we could not complete a full sweep for all model sizes (e.g. we train our 2.5B model up to 500 tokens/parameter). Our dataset consists of trillions of tokens of general web text and code. For all experiments, we train for only a single epoch and do not repeat data. Further details are provided in Section \ref{sec:traindetails} in the Appendix.

Furthermore, to ensure that loss is a good proxy for downstream metrics, we evaluate each model on a version of the open source Evaluation Gauntlet \cite{MosaicML2023LLMEvaluation}, with tasks in five categories:
\begin{itemize}
\item \textbf{World Knowledge}: This includes Jeopardy \cite{jeopardy}, MMLU \cite{hendrycks2020measuring}, BIG-bench WikiData \cite{srivastava2022beyond}, ARC Easy and ARC Challenge \cite{arc}.
\item \textbf{Commonsense Reasoning}:  BIG-bench Strategy QA and BIG-bench Strange Stories  \cite{srivastava2022beyond}, COPA \cite{copa}, PIQA \cite{piqa}, OpenBook QA \cite{openbook_qa}, Common Sense QA \cite{talmor2018commonsenseqa}, and SIQA \cite{sap2019socialiqa}.
\item \textbf{Reading Comprehension}: SQuAD \cite{squad}, BoolQ \cite{clark2019boolq}, CoQA \cite{reddy-etal-2019-coqa}, and AGI Eval \cite{zhong2023agieval}.
\item \textbf{Language Understanding}: LAMBADA \cite{paperno2016lambada}, HellaSwag \cite{zellers2019hellaswag}, Winograd Schema Challenge \cite{winograd}, and Winogrande \cite{winogrande}.
\item \textbf{Symbolic Problem Solving}: BIG-bench Elementary Math QA \cite{srivastava2022beyond} BIG-bench Dyck Languages \cite{srivastava2022beyond}, BIG-bench Operators \cite{srivastava2022beyond}, Math QA \cite{math_qa}, LogiQA, GSM8k \cite{cobbe2021gsm8k}, SVAMP \cite{patel-etal-2021-nlp}, AGI Eval SAT Math and AGI Eval LSAT \cite{zhong2023agieval}.
\end{itemize}
To compute model performance on the above datasets, the Evaluation Gauntlet uses the \href{https://docs.mosaicml.com/projects/composer/en/stable/api_reference/generated/composer.metrics.InContextLearningQAAccuracy.html}{InContextLearningQAAccuracy} for question-answering tasks, \href{https://docs.mosaicml.com/projects/composer/en/stable/api_reference/generated/composer.metrics.InContextLearningLMAccuracy.html}{InContextLearningLMAccuracy} for language-modeling tasks, and \href{https://docs.mosaicml.com/projects/composer/en/stable/api_reference/generated/composer.metrics.InContextLearningMultipleChoiceAccuracy.html}{InContextLearningMultipleChoiceAccuracy} for multiple-choice tasks. All metrics are from the Composer library \cite{composer}.

The Gauntlet Average is an average of the accuracy of all the above tasks, with each task weighted equally after subtracting out the task's baseline (random) accuracy and normalizing.

\section{Results}
\label{sec:results}
In Figure \ref{fig:experimentresults}, we show the major results of our experiments. Our key finding is that loss continues to decrease as we increase tokens per parameter, even to extreme ratios. Although it takes exponentially more tokens to reduce loss at large ratios, loss does not plateau as we scale to 10,000 tokens per parameter for our 150M model. For models larger than 150M, we only test up to a maximum of 1,000 tokens per parameter due to resource constraints, and at this scale we also see no evidence of loss flat-lining. Further experiments are needed to see how loss scales beyond this point. 

Existing literature claims there exists a critical model size, below which we cannot match a Chinchilla-optimal model's quality \citep{blogpost}. 
Since our experiments do not see any plateauing of loss as we scale token ratios, we find no evidence to support the critical size hypothesis, although further testing is needed at extreme scales---it is possible that beyond 10,000 tokens/parameter, behavior changes.

We also show results for downstream metrics. Fig. \ref{fig:gauntlet_tpr} shows that the Gauntlet Average improves as we increase tokens per parameter --- again, we find no evidence of a ``saturation" point, beyond which additional tokens do not result in better performance. In fact, as Fig. \ref{fig:gauntlet_loss} shows, as loss decreases, smaller decreases in loss lead to larger improvements in downstream accuracy.

In Fig. \ref{fig:gauntlet_loss}, we plot the Gauntlet Average as a function of loss. Loss and Gauntlet Average are tightly correlated, showing that improvements in loss are excellent predictors for improvements in general model quality. Our results show that practitioners interested in predicting downstream metrics as a function of model parameters and token counts can make use of existing scaling laws to accurately understand how their downstream metrics change at scale.  

\begin{figure}
    \centering
    \includegraphics[width=1.0\linewidth]{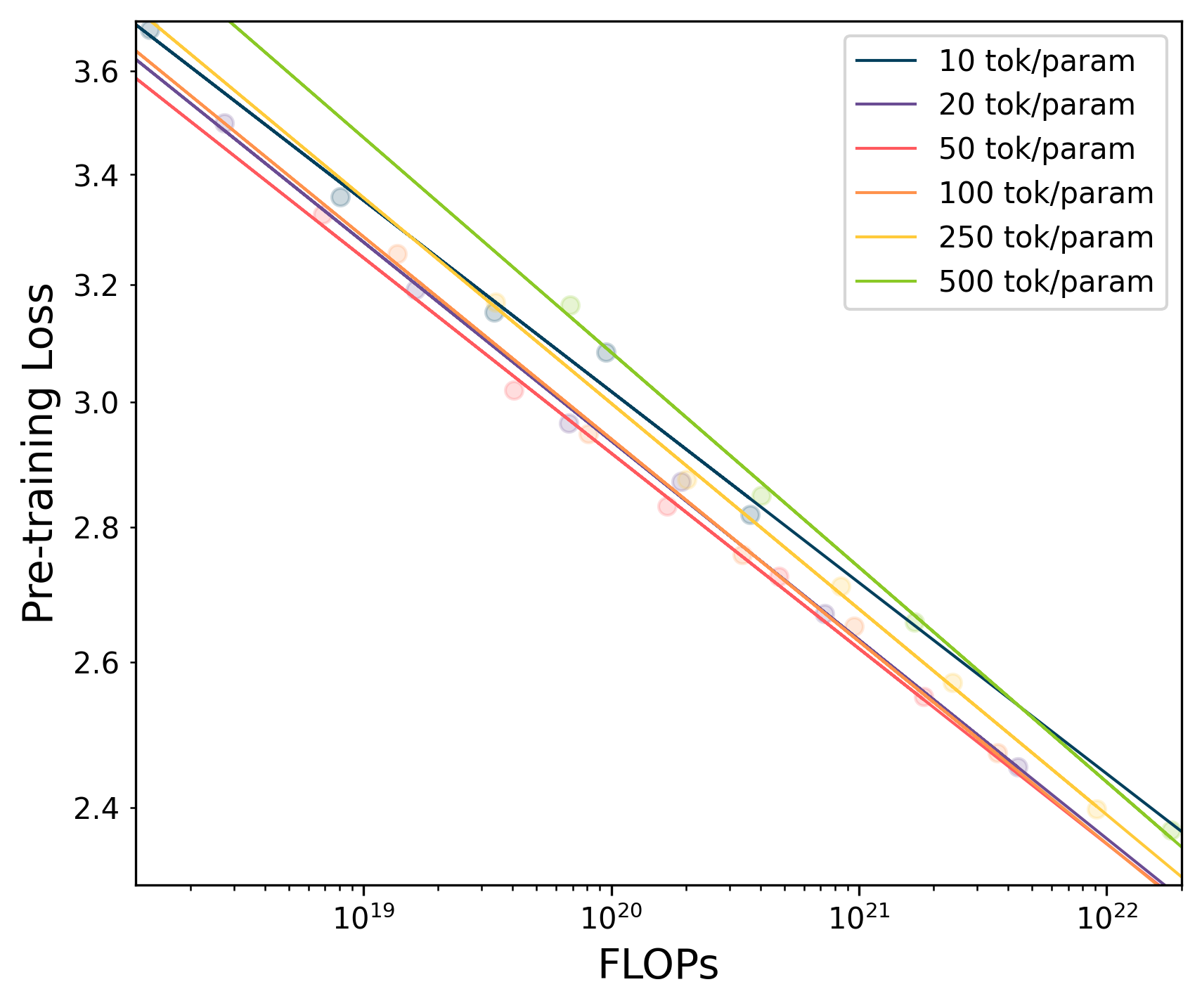}
    \caption{Log-log plot of loss vs. FLOPs, grouped by token/parameter ratio. The slope of each trendline shows how efficiently models learn at each ratio.}
    \label{fig:loglogplot}
\end{figure}

In Figure \ref{fig:loglogplot}, we plot loss vs. FLOPs, grouping our data points by token-per-parameter ratio (up to 500 tokens/param) instead of model size, following \citet{gadre2024language}. For token-per-parameter ratios $\geq 20$, our lines of best fit are nearly parallel, indicating that models learn with similar efficiency. This result is striking, as it indicates that loss decreases at similar rates for each additional FLOP regardless of whether training occurs in the standard Chinchilla regime or at extreme ratios (500 tokens/param). Our results also confirm that training at low ($<20$) token-per-parameter ratios is less computationally efficient, in line with the results from the Chinchilla paper.

In Section \ref{sec:furtherresults} in the Appendix, we show results for each Gauntlet category. Since they consist of fewer tasks, category averages show less consistent correlation with data and model sizes than the overall Gauntlet Average, because on many tasks smaller models are not able to achieve significantly better performance than a random baseline \cite{Barton2024}.

\section{Parametric Fitting}
\begin{figure}
    \centering
    \includegraphics[width=1.0\linewidth]{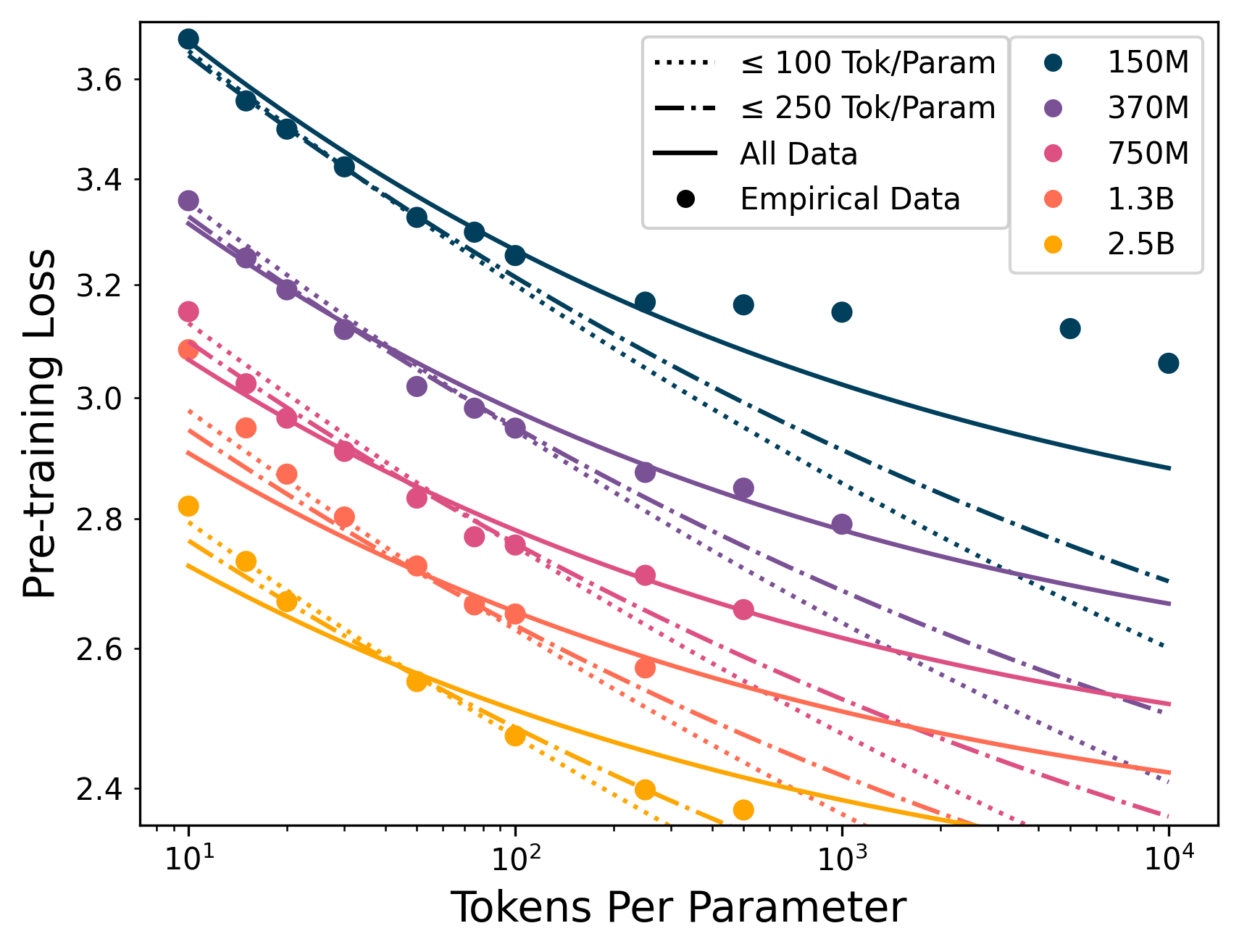}
    \caption{Chinchilla parameteric scaling curves fit to progressively larger subsets of our empirical training data. The coefficients that define these curves are detailed in Table \ref{table:parametricfits}.}
    \label{fig:curves}
\end{figure}
\label{sec:parametric}
The Chinchilla scaling law has five coefficients ($A, B, E, \alpha$ and $\beta$) which were found empirically by fitting Eq. \ref{eq:loss} to data collected from 400 training runs by \citet{chinchilla}. These coefficients determine precisely how much the scaling law weighs increasing model parameters vs. increasing training data.
\citet{dataconstrained} found similar coefficient values after training 54 similarly-sized runs on C4, one of the datasets tested in the Chinchilla paper. In both cases, the empirical data collected was largely from small token-to-parameter training runs ($<$\textapprox100). Since \citet{chinchilla} and \citet{dataconstrained} focused on optimizing training compute and data efficiency, respectively, it is reasonable that they concentrated empirical data collection around the training efficiency ``sweet spot," around 20 tokens per parameter. However, when we incorporate inference costs into the picture, we may need to train models for hundreds or thousands of tokens per parameter. Thus, we require scaling laws that can accurately predict training loss far beyond typical data ratios.

To understand how well the Chinchilla parametric fitting procedure generalizes to loss values for extremely long training runs, we fit parametric curves on successively larger subsets of our training data. First, we use only training runs with no more than 100 tokens per parameter, followed by runs with no more than 250 tokens per parameter, etc. For each of the curves, we follow the procedure described in the Sec. D.2 of the Chinchilla paper, minimizing the objective in Eq. \ref{eq:huber}, where LSE is the log-sum-exp operator, and $\delta=10^{-3}$. We use the L-BFGS algorithm to perform the minimization, initialized from a grid of starting points for each of the parameters. We then set $A, B, E = \exp(a), \exp(b), \exp(e)$ and minimize the Huber loss in the following equation:
\begin{align}
\label{eq:huber}
\begin{split}
\min_{a, b, e, \alpha, \beta} \sum_{\text{Run } i} \text{Huber}_{\delta}\Big(\text{LSE}\big[& a - \alpha \log (N^{(i)}),\\ 
& b - \beta \log(\Dtr^{(i)}), e\big] - \log(L^{(i)}) \Big)
\end{split}
\end{align}
Our results are shown in Table \ref{table:parametricfits} and visualized in Figure \ref{fig:curves}. First, we see that our parameteric curves follow a consistent trend---as we use more extreme data in our fitting procedure, our scaling curves become flatter. This trend suggests that if we only use data from typical token ratios to determine our scaling law coefficients, we will \textit{overestimate} the impact of additional training data as we move towards the long-data-ratio regime. Taken together with our results from Sec. \ref{sec:results}, our experiments show that although models continue to learn at extreme training durations, they do so more slowly than scaling laws predict.  

Second, somewhat surprisingly, none of of our parametric curves fit our 150M long-ratio training results well. It appears that as we extend training duration far beyond typical Chinchilla ratios, the parametric loss function is \textit{not flexible enough} to accurately model the behavior of both smaller ($\leq$150M) and larger models. Further research is needed to understand the limits of scaling law extrapolation.

We note that our experimentally-derived coefficients differ from those found in the Chinchilla paper. This is expected, as both the data and the model architectures we used to train all checkpoints are quite different from the original work \cite{chinchilla}. \citet{besiroglu2024chinchilla} noted in a replication study that the coefficients reported in \citet{chinchilla} were rounded in ways that leads to significant bias in scaling law predictions.

\begin{table}
  \caption{Parametric fitting on data including extreme training runs. The Chinchilla coefficients from \citet{chinchilla} are included for reference.}
  \label{table:parametricfits}
  \centering
  \begin{tabular}{lccccc}
    \toprule
    Data & $\alpha$ & $\beta$ & A & B & E \\
    \midrule
    $\leq$100 tok/param & 0.08 & 0.13 & 7.199 & 25.97 & 0.17 \\
    $\leq$250 tok/param & 0.13 & 0.16 & 14.23 & 39.54 & 0.98 \\
    $\leq$500 tok/param & 0.13 & 0.16 & 17.07 & 35.80 & 0.95 \\
    All Data & 0.18 & 0.24 & 33.66 & 138.9 & 1.45\\
    \midrule
    Chinchilla & 0.34 & 0.28 & 406.4 &  410.7 & 1.69 \\
    \bottomrule
  \end{tabular}
\end{table}

Since \citet{chinchilla} trained almost exclusively on small-duration training runs ($\leq$\textapprox100 tokens/parameter), our results suggest that that in their original form, the Chinchilla scaling laws do not extend to extreme-duration training runs. To the extent that they do, current scaling laws overestimate the improvements in loss that stem from long-duration training on additional data.

These results have significant impact as the field continues to train longer and longer. For example, LLaMA 7B was trained on 1 trillion tokens \cite{llama}, Llama 2 7B was trained on 2 trillion tokens \cite{llama2}, and Llama 3 8B was trained on 15 trillion tokens \cite{llama3}. How much of the quality difference between these models is due to simply training on more data? How much is due to other changes, like architecture modifications or data quality improvements? Our results indicate that if we apply the Chinchilla scaling laws to understand the scaling component of this quality improvement, we will likely overestimate the its impact of more data versus the other changes.

Further work is needed to fully characterize scaling laws at extreme token to parameter ratios. Due to resource constraints, we do not collect data at the same scale as the Chinchilla paper---both in terms of model size (we only test up to 6B vs. 16B), and number of training runs (47 vs. 400). 
Little research has been conducted to explore the minimum data required to accurately fit scaling laws coefficients (see \citet{besiroglu2024chinchilla}).
\begin{figure*}
        \begin{subfigure}[b]{0.33\textwidth}
                \centering
                \includegraphics[width=1.00\linewidth]{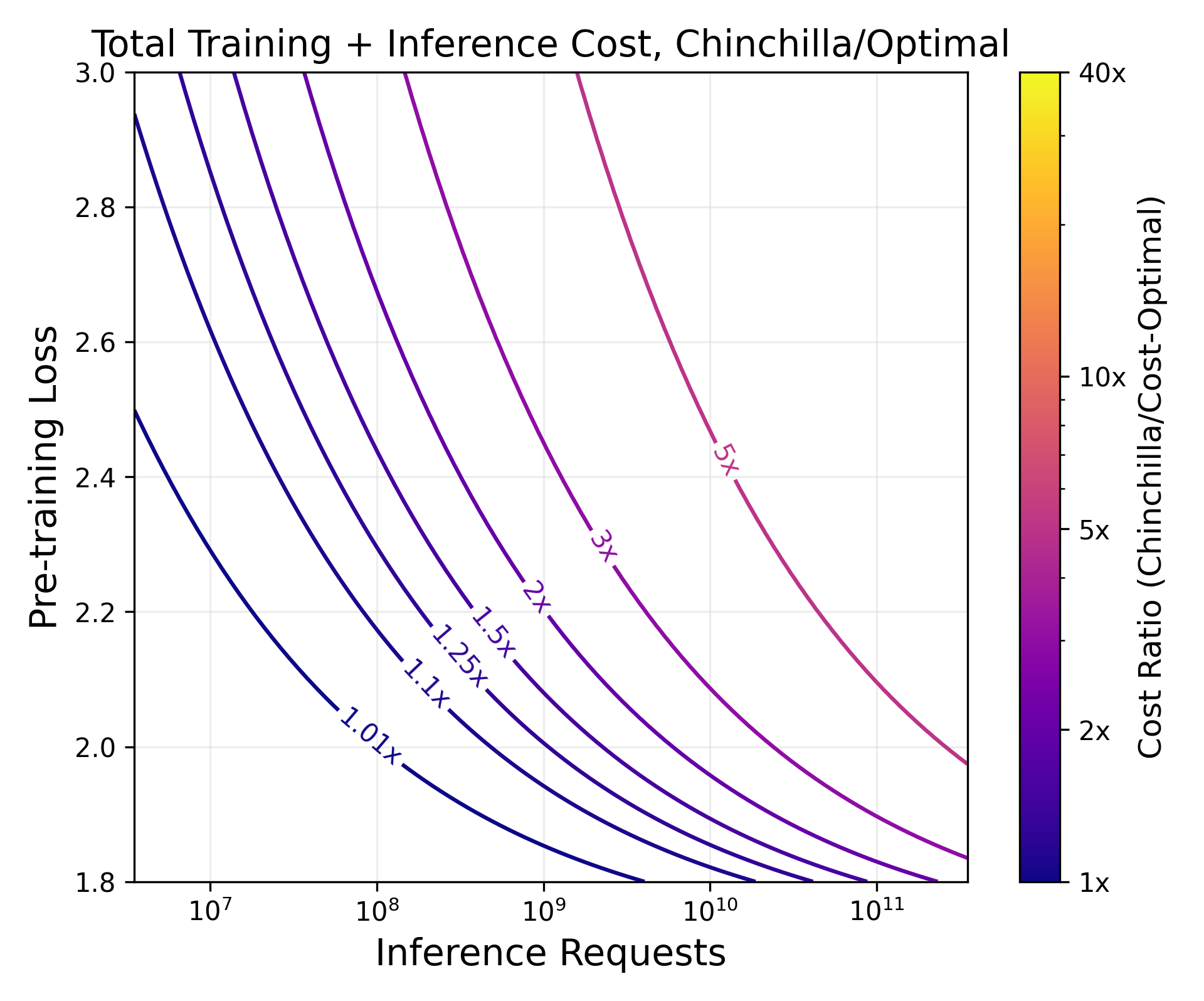}
                \caption{}
                \label{fig:costratio}
        \end{subfigure}%
        \begin{subfigure}[b]{0.33\textwidth}
                \centering
                \includegraphics[width=1.00\linewidth]{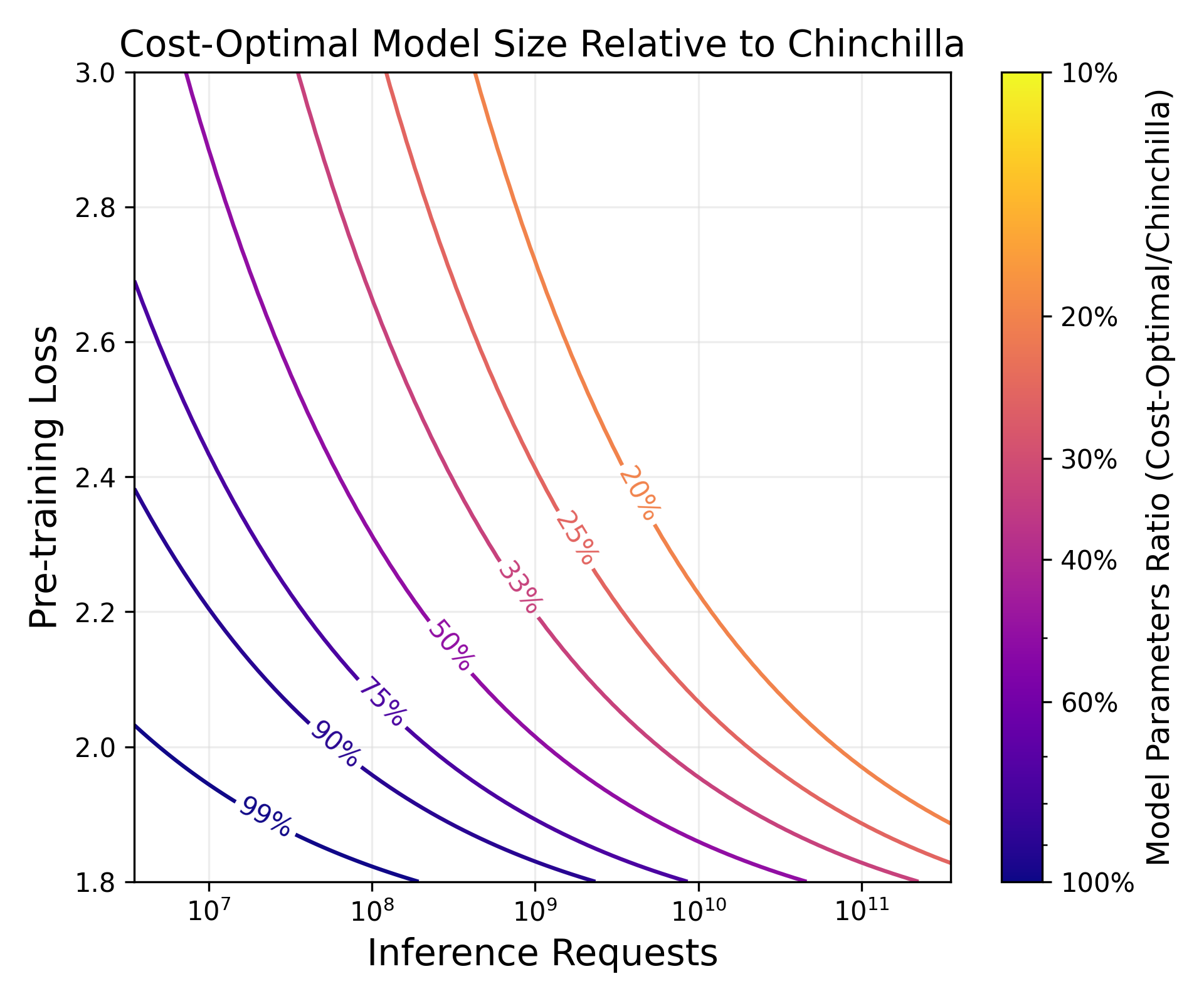}
                \caption{}
                \label{fig:costparams}
        \end{subfigure}%
        \begin{subfigure}[b]{0.33\textwidth}
                \centering
                \includegraphics[width=1.00\linewidth]{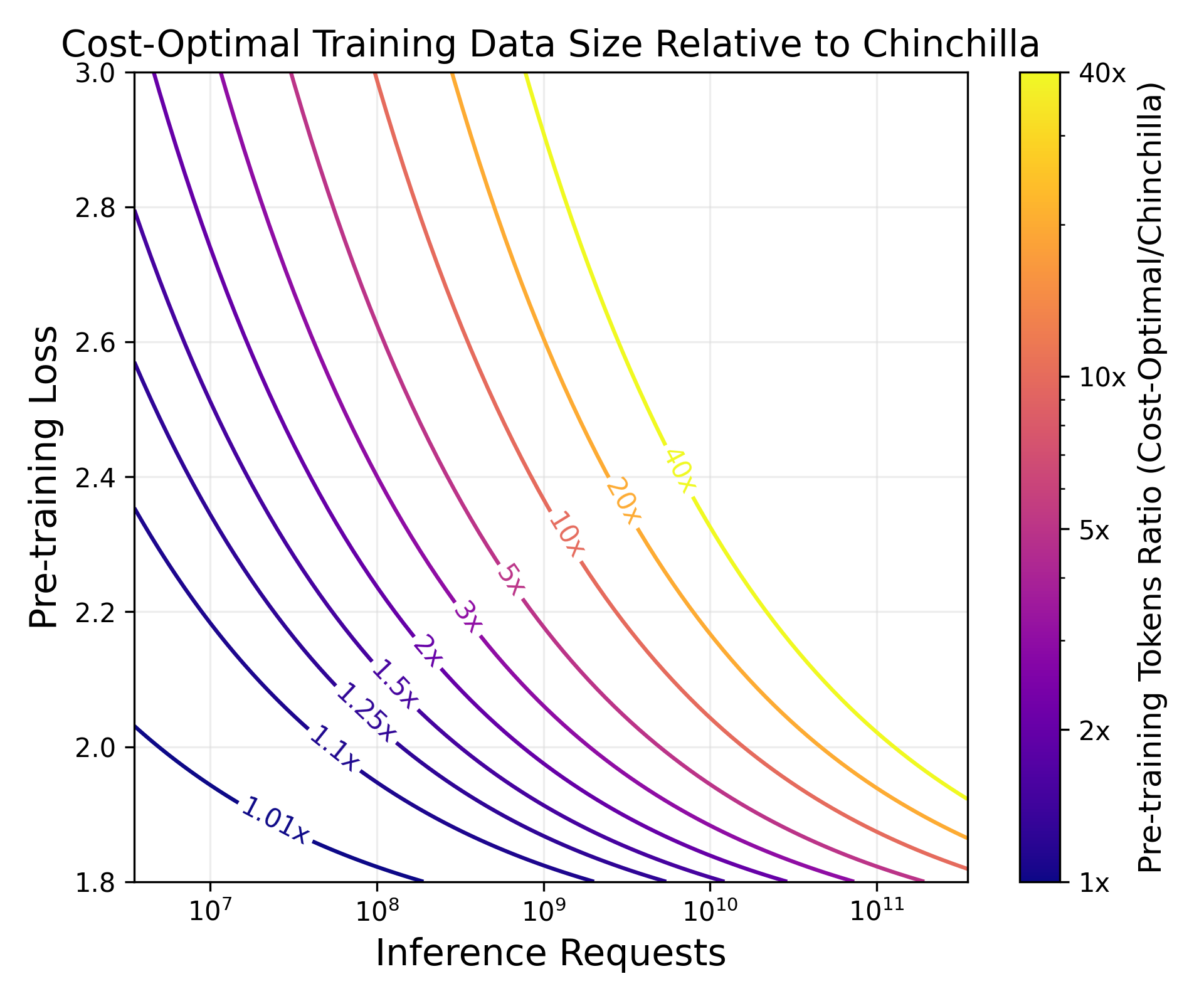}
                \caption{}
                \label{fig:costtokens}
        \end{subfigure}%
        \caption{Ratios of (a) total training + inference cost, (b) model parameters, and (c) pre-training tokens, for cost-optimal models via our real-world estimation method vs. Chinchilla-style models. Results in this figure are shown with the following settings: training with 50\% MFU, inference input with 50\% MFU, generation with 1\% MFU. Inference requests have 70 input tokens and 215 output tokens each, aligning with averages from real-world data \citep{lmsys}. To mimic a realistic scenario, we calculate costs assuming training occurs on A100-80GB and inference occurs on A100-40GB accelerators after INT8 quantization (see Sec. \ref{sec:pricing} for details).}
        \label{fig:cost}
\end{figure*}
\section{Estimating Real-World Cost Optimality}
\label{sec:costopt}
Our method introduced in Section \ref{sec:compopt} optimizes purely for minimum total (training + inference) FLOPs. However, this has significant drawbacks which limit its applicability to real-world deployments. The real-world cost of an inference request of $3D$ tokens is generally different than the cost to train on $D$ tokens. For instance, inference hardware utilization can be much lower than training utilization, since small batch size computation can result in low Model FLOPs Utilization (MFU). MFU can be as low as \textapprox1\% for inference \citep{palminference} but is typically 40-60\% during training \citep{megatronlm}. Utilization is also different for input tokens vs. output tokens --- since input tokens (prompts) are typically processed in a single forward pass, utilization is typically near training levels. By contrast, during generation, output tokens must be produced sequentially, resulting in low utilization due to memory bandwidth constraints. Another complicating factor is that inference operations can sometimes be cheaper than training FLOPs, since models can be quantized before inference time, turning 16- or 32-bit floating-point operations into 4- or 8-bit integer operations which run more efficiently on the same hardware. Quantization can also enable LLMs to fit on GPUs with less VRAM, so training and inference may occur on different hardware altogether \citep{gptq}. 

To estimate the real-world cost of inference, we modify Eq. \ref{eq:1} to account for hardware utilization: $\Utr$, $\Uinp$, and $\Uout$ are our training, inference input, and inference output MFUs, respectively. In addition, we add parameters for training and inference cost per FLOP, $C_\text{tr}$ and $C_\text{inf}$.  Our new objective is:
\begin{align}
\label{eq:3}
\begin{split}
N^*(\ell, \Dinp, \Dout), &\Dtropt(\ell,\Dinp, \Dout) =\\
&\argmin_{N, D \mid L(N, \Dtr) = \ell} \Bigg[\frac{\Ctr}{\Utr} \Tf(N, \Dtr)\\
&+ \sum_{i}^{} \frac{\Cinf}{\Uinp}\If(N, D_{\text{inp}}^{(i)})\\
&+ \sum_{i}^{} \frac{\Cinf}{\Uout}\If(N, D_{\text{out}}^{(i)})\Bigg].
\end{split}
\end{align}
We again use the approximations for FLOPs for transformer models, reducing the above equation to:
\begin{align}
\label{eq:4}
\begin{split}
&N^*(\ell, \Dinp, \Dout), D_{\text{tr}}^{*}(\ell, \Dinp, \Dout) =\\
&\argmin_{N, \Dtr \mid L(N, \Dtr) = \ell}\frac{6 N \Dtr\Ctr}{\Utr} + 2N\Cinf\bigg[\frac{\Dinp}{\Uinp} + \frac{\Dout}{\Uout}\bigg]
\end{split}
\end{align}
Eq. \ref{eq:4} is a simplified model of real-world costs: we leave aside latency requirements and assume MFU and cost per FLOP do not depend on model size, configuration, or sequence length. Still, our approximation is flexible enough to account for heterogeneous hardware utilization and costs.

In Figure \ref{fig:cost}, we show how inference-adjusted cost-optimal models compare to Chinchilla-style models, assuming typical training and inference hardware costs and MFU. For a 30B-Chinchilla-quality model, LLM practitioners expecting 1.5B inference requests can reduce costs by 17$\%$ by instead training a 16B model on 3.35T tokens.
In Sec. \ref{sec:costoptappendix}, we show further results for various configurations.  

Comparing our compute-optimal analysis in Fig. \ref{fig:ratios} to our real-world cost analysis in Fig. \ref{fig:cost}, we see that for the same inference demand of 2T tokens (7.02B requests), a Chinchilla-70B model requires only \textbf{1.3\%} extra FLOPs compared to an equal-quality \textit{compute-optimal} model, but costs \textbf{36\%} more than a \textit{cost-optimal} model. This difference is attributable to the 50$\times$ lower MFU of each inference output token compared to training, which our FLOP-based analysis in Sec. \ref{sec:compopt} fails to capture.

\section{Related Work}

Many studies have contributed to the development of scaling laws for LLMs, including \citet{hestness2017deep, hestness2019beyond},  \citet{rosenfeld2019constructive}, \citet{henighan2020scaling}, \citet{scalinglaws}, \citet{sorscher2022beyond}, \citet{tay2022scaling}, and \citet{caballero2022broken} (see \citet{villalobos2023scaling} for a review). Some of these studies focused on scaling laws for transfer settings (i.e. downstream performance), such as \citet{hernandez2021scaling,mikami2021scaling,abnar2021exploring} and \citet{tay2022scaling}.

A few studies have also (gently) critiqued the general parametric function fitting approach of \citet{chinchilla}. \citet{besiroglu2024chinchilla} attempted to replicate the methodology used in \citet{chinchilla} and found that the confidence intervals reported in the original study were implausibly narrow. The broader implication is that confidence intervals are quite wide for parametric function fitting on a small number of data points. This is also a potentially valid critique of our empirical results, as our analysis only includes 47 separate experiments (Chinchilla included more than 400 experiments). In a similar vein, \citet{porian2024resolving} investigate discrepancies between the scaling laws from \citet{scalinglaws} and \citet{chinchilla}.

A handful of compelling scaling law papers have been published since 2023, when an earlier version of this work was first presented \cite{sardana2023beyond}. For example, \citet{krajewski2024scaling}  characterize differences in scaling properties between dense transformers and Mixture of Expert (MoE) models. More theoretical studies include \citet{michaud2024quantization} and \citet{bordelon2024dynamical}.
\citet{paquette20244+} uses phase-plane analysis to characterize scaling laws in the compute-limited, infinite-data regime. 
\citet{ruan2024observational} builds scaling laws from multiple LLM model families using low-dimensionality tools applied to publicly available data of LLM performance on open benchmarks.

The results presented in \citet{gadre2024language} are particularly relevant to this paper. The authors train 100 models between the sizes of 1.4B and 6.9B parameters and on data with tokens-per-parameter ratios between 20 and 640. Similar to our study, they find reliable scaling laws in these model and data regimes. They also find that downstream task performance is strongly correlated to LLM perplexity.

\section{Conclusion}
In this work, we modify the Chinchilla scaling laws to account for both the computational and real-world costs of inference. As inference demand approaches pre-training data size, the additional cost pushes the optimal tokens-to-parameters ratio towards smaller and longer-trained models.

We experimentally validate the hypothesis that very small models, trained on enough data, can match larger Chinchilla-style ones. Both in terms of loss and downstream metrics, model quality improves as tokens per parameter increase. Further work is needed to show if this scales beyond 10,000 tokens per parameter, or at larger model sizes. Still, our results show that practitioners operating in inference-heavy regimes, or with limited deployment memory, can scale training duration considerably longer than current literature suggests and see quality improvements.

Finally, we show evidence that the Chinchilla parametric coefficient fitting procedure overestimates the reduction from additional training data when applied to extremely data-heavy training runs. More work is needed to develop scaling laws that apply precisely at a wide range of ratios.

\section*{Acknowledgements}
We thank Daya Khudia for his support and Mihir Patel and Linden Li for their feedback on this manuscript.

\section*{Impact Statement}
This paper presents work whose goal is to reduce the costs of producing large language models. LLMs have the potential to produce factual, high-quality text, but also biased or harmful outputs. By reducing the costs train capable LLMs, we make them more accessible to scientific researchers, industry, and the general population.

\bibliography{main}
\bibliographystyle{icml2024}

\newpage
\appendix
\onecolumn
\section{No Analytic Solution for Inference-Compute Optimality} \label{sec:proof1}
In this section, we prove that it is not possible to analytically derive the optimal model size and pre-training token count according to the third Chinchilla law, after accounting for the computational cost of inference. Conditioned on model quality, we assume that inference demand does not depend on model size and can be estimated prior to training.

\begin{theorem}
\label{noanalytic}
Given a fixed model quality and inference demand, there exists no general analytic solution for the compute-optimal model size and pre-training token count according to the third Chinchilla law, after accounting for the computational cost of inference. 
\end{theorem}
\begin{proof}
By Eq. \ref{eq:2}, the overall compute cost in FLOPs for training a model with $N$ parameters on $\Dtr$ tokens and running inference on $\Dinfi$ tokens is given by $C(N, \Dtr, \Dinfi) = 6N\Dtr + 2N\Dinfi$. 

We seek the minimum overall compute budget to train and deploy a model of a given quality and inference demand. Formally, we optimize the objective:
\begin{align}
\min C(N, \Dtr, \Dinfi)
\end{align}
subject to the constraint
$L(N, \Dtr) =  E + \frac{A}{N^\alpha} + \frac{B}{D_{\text{tr}}^\beta} = \ell$.

This constraint, from the third Chinchilla law, ensures we are minimizing compute while fixing model quality (pre-training loss). $A = 406.4, B = 410.7, E = 1.69, \alpha = 0.336$, and $\beta = 0.283$ are constants determined empirically by \citet{chinchilla}.\footnote{The Chinchilla paper reports $\alpha = 0.34$ and $\beta = 0.28$. However, these are rounded values; to better fit the results reported in Table A.3 of \citet{chinchilla}, we use $\alpha = 0.336$ and $\beta = 0.283$, as in \citet{blogpost}.}

We solve this optimization problem via the method of Lagrange multipliers. The gradients are:
\begin{align}
\nabla C(N, \Dtr) = (6D + 2\Dinfi) \hat{i} + 6N \hat{j}
\end{align}
\begin{align}
\nabla L(N, \Dtr) = -\alpha A N^{-\alpha - 1}\hat{i} - \beta B \Dtr ^{-\beta - 1} \hat{j}
\end{align}
We have three equations and three variables ($\Dtr, N, \lambda$), where $\lambda$ is our Lagrange multiplier:
\begin{align}
6\Dtr + 2\Dinfi  = -\lambda \alpha A N^{-\alpha - 1} &&
6N = -\lambda \beta B \Dtr ^{-\beta - 1}
&&
E + \frac{A}{N^\alpha} + \frac{B}{\Dtr^{\beta}} = \ell
\end{align}
With some algebraic manipulation, we can eliminate $\lambda$ and write $\frac{A}{N^\alpha}$ in terms of $\Dtr$:
\begin{align}
\frac{A}{N^{\alpha}} = \frac{3\beta B \Dtr^{-\beta} + \Dinfi \beta B \Dtr^{-\beta-1}}{3\alpha}    
\end{align}
We are left to solve the following equation for $\Dtr$: 
\begin{align}\label{eq:11}
0 = (E - \ell) + \Big[\frac{\beta B}{\alpha} + B\Big] \Dtr^{-\beta} + \frac{\Dinfi \beta B}{3\alpha}\Dtr^{-\beta-1}
\end{align}
Thus, determining $\Dtr$ as a function of $\Dinfi$ and $\ell$ involves finding the roots of equations of the form $ ax^{-1.283} + 756.6x^{-0.283} + c = 0$ for arbitrary $a$ and $c > 0$, which is not possible in general.

The best-fit constants $A, B, E, \alpha$ and $\beta$ vary based on the exact dataset and model architecture. Outside of a handful of special-case values of $\beta$ for which Eq. \ref{eq:11} can be manipulated into a low-degree polynomial, it is intractable.  
\end{proof}

\section{Further Results}
\subsection{Compute-Optimal Results}\label{sec:compoptappendix}
We present further results from our analysis in Sec. \ref{sec:compopt}. In Table \ref{table:computetable}, we show the computational cost (in FLOPs) to train and run inference for Chinchilla-style models of various sizes and inference demands. We then calculate the \textit{compute-optimal} model configuration to reach the same quality (equal loss) and run inference, and note the overall compute reduction.

\begin{table}
  \caption{Compute-Optimal vs. Chinchilla-style Models for Selected Configurations.}
  \label{table:computetable}
  \centering
  \begin{tabular}{llccccccr}
    \toprule
    & & & & & & & \\
    & & \multicolumn{3}{c}{Chinchilla Model} & \multicolumn{3}{c}{Compute-Optimal Model} \\
    \cmidrule(r){3-5} \cmidrule(r){6-8}
    Inference & Train & & Training &  &  & Training &  & FLOP
    \\ Tokens & Loss & Params & Tokens & FLOPs & Params & Tokens & FLOPs & Reduction\\
    \midrule
    50B & 2.53 & 1B & 27.4B  & 2.64e20 & 6.33M  & 46.8B & 2.41e20 & 9.1\%\\
    200B & 2.13 & 7B & 276B &  1.44e22   & 5.4B & 367B & 1.40e22 & 2.6\%    \\
    1T & 2.05 & 13B & 577B & 7.10e22 & 8.32B & 967B & 6.49e22 & 8.5\% \\
    5T & 1.96 & 30B & 1.56T  & 5.80e23 & 16.4B & 3.27T & 4.86e23 & 16\% \\
    10T & 1.89 & 70B     & 4.26T & 3.19e24 & 41.6B & 7.92T & 2.81e24 & 12\% \\
    \bottomrule
  \end{tabular}
\end{table}

\subsection{Cost-Optimal Results}\label{sec:costoptappendix}
We show additional results from our cost-optimality analysis in Sec. \ref{sec:costopt}. In Table \ref{table:costtable}, we show the total training plus inference costs for Chinchilla models of various sizes at different levels of inference demands. We then calculate costs for equivalent-quality (i.e. same pre-training loss) \textit{cost-optimal} models and show the overall savings. We use the same settings from Figure \ref{fig:cost}, designed to mimic a typical real-world deployment: training and inference input at 50\% MFU, generation at 1\% MFU \cite{megatronlm, palminference}. Each inference request has 70 input tokens and 215 output tokens, in accordance with averages from the 
LMSYS-Chat dataset of 1M inference requests from \citet{lmsys}. Costs are calculated assuming training and inference on A100-80GB and A100-40GB accelerators, respectively. We further assume the model parameters are quantized to eight-bit integers prior to inference, which is commonly done with no quality reduction \cite{smoothquant}. All costs are reported in US dollars.

\begin{table}
  \caption{Cost-Optimal vs. Chinchilla-style Models for Selected Configurations.}
  \label{table:costtable}
  \centering
  \begin{tabular}{llccccccr}
    \toprule
    & & & & & & & \\
    & & \multicolumn{3}{c}{Chinchilla Model} & \multicolumn{3}{c}{Cost-Optimal Model} \\
    \cmidrule(r){3-5} \cmidrule(r){6-8}
    Inference & Train & & Training & Total  &  & Training & Total  & Cost
    \\ Requests & Loss & Params & Tokens & Cost (\$) & Params & Tokens & Cost (\$) & Savings\\
    \midrule
    175M & 2.53 & 1B & 27.4B  & 3.77K & 327M  & 152B &  1.89K & 50\% \\
    702M & 2.13 & 7B & 276B & 124K     & 2.90B & 929B  & 81.8K & 34\%    \\
    3.51B & 2.05 & 13B & 577B & 987K & 430B & 3.1T & 500K & 49\% \\
    17.5B & 1.96 & 30B & 1.56T & 10.8M & 8.58B & 12.1T & 4.52M & 58\% \\
    35.1B & 1.89 & 70B & 4.26T & 51.5M & 21.5B & 27T & 23.8M & 54\% \\
    \bottomrule
  \end{tabular}
\end{table}

\subsection{GPU Details}\label{sec:pricing}
GPU pricing varies based on vendor and fluctuates over time. At the time of writing, an A100-40GB costs USD \$1.10/hr and an A100-80GB costs \$1.50/hr on Lambda Labs \cite{lambdalabs}. We use these values in our cost analysis in Sec. \ref{sec:costopt} and in Table \ref{table:costtable}. Both variants have a peak performance of $3.12 \times 10^{14}$ dense FP16/BF16 operations and  $6.24 \times 10^{14}$ INT8 operations per second \cite{a100datasheet}.

\section{Model Training}\label{sec:traindetails}
We train MPT-style transformer models \cite{mpt} ranging in size from 150M to 6B parameters, on token/parameter ratios from 10 to 10,000. In Table \ref{table:traintable}, we provide training configuration details. All models were trained with ALiBi \cite{alibi}, Grouped Query Attention \cite{gqa}, the Lion optimizer \cite{lion} ($\beta_1 = 0.9, \beta_2 = 0.95$) with weight decay equal to the learning rate, cosine warmup ($\alpha_f = 0.1$) with a duration equal to 3 times the number of model parameters, and norm gradient clipping (threshold = 1). A maximum sequence length of 4096 tokens was used. 
A smaller batch size was used for smaller models so that low-token-count training runs see enough update steps to learn properly. For all experimental results in this work, we use the smoothed final training loss over the last ten batches to reduce noise from minor batch-level variations.

In the rightmost column of Table \ref{table:traintable}, we list our experiment training token ratios. Note that for the larger models, our training durations are limited by our computational resources.

\begin{table}
  \caption{Model Training Configurations.}
  \label{table:traintable}
  \centering
  \begin{tabular}{llccccccr}
    \toprule
    Name & Params & d\_model & n\_heads & n\_layers & Learning Rate & Batch Size & Tokens/Parameter \\
    \midrule
    150M & 151M & 768 & 12  & 12 & 4.603e-4 & 160\tablefootnote{The 10,000 token/parameter 150M model was trained with batch size 960 due to compute issues.} & 10,15,20,30,50,75,100, & \\
    & & & & & & & 250,500,1000,5000,10000 \\
    370M & 367M & 1024 & 16 & 24 & 3.453e-4 & 320 & 10,15,20,30,50,75,100, \\
     & & & & & & &250,500,1000 \\
    750M & 749M & 1536 & 12 & 24 & 2.302e-4 & 480 & 10,15,20,30,50,75,100,250,500 \\
    1.3B & 1.26B & 2048 & 16 & 24 & 1.726e-4 & 960 & 10,15,20,30,50,75,100,250\\
    2.5B & 2.46B & 2560 & 20 & 32 & 1.381e-4 & 960 & 10,15,20,50,100,250,500 \\
    6B &  6.05B & 4096 & 32 & 32 & 8.632e-5 & 960 & 20 \\
    \bottomrule
  \end{tabular}
\end{table}

\section{Experiment Results}
\label{sec:furtherresults}
\begin{figure}
        \begin{subfigure}[b]{0.33\textwidth}
                \centering
                \includegraphics[width=1.00\linewidth]{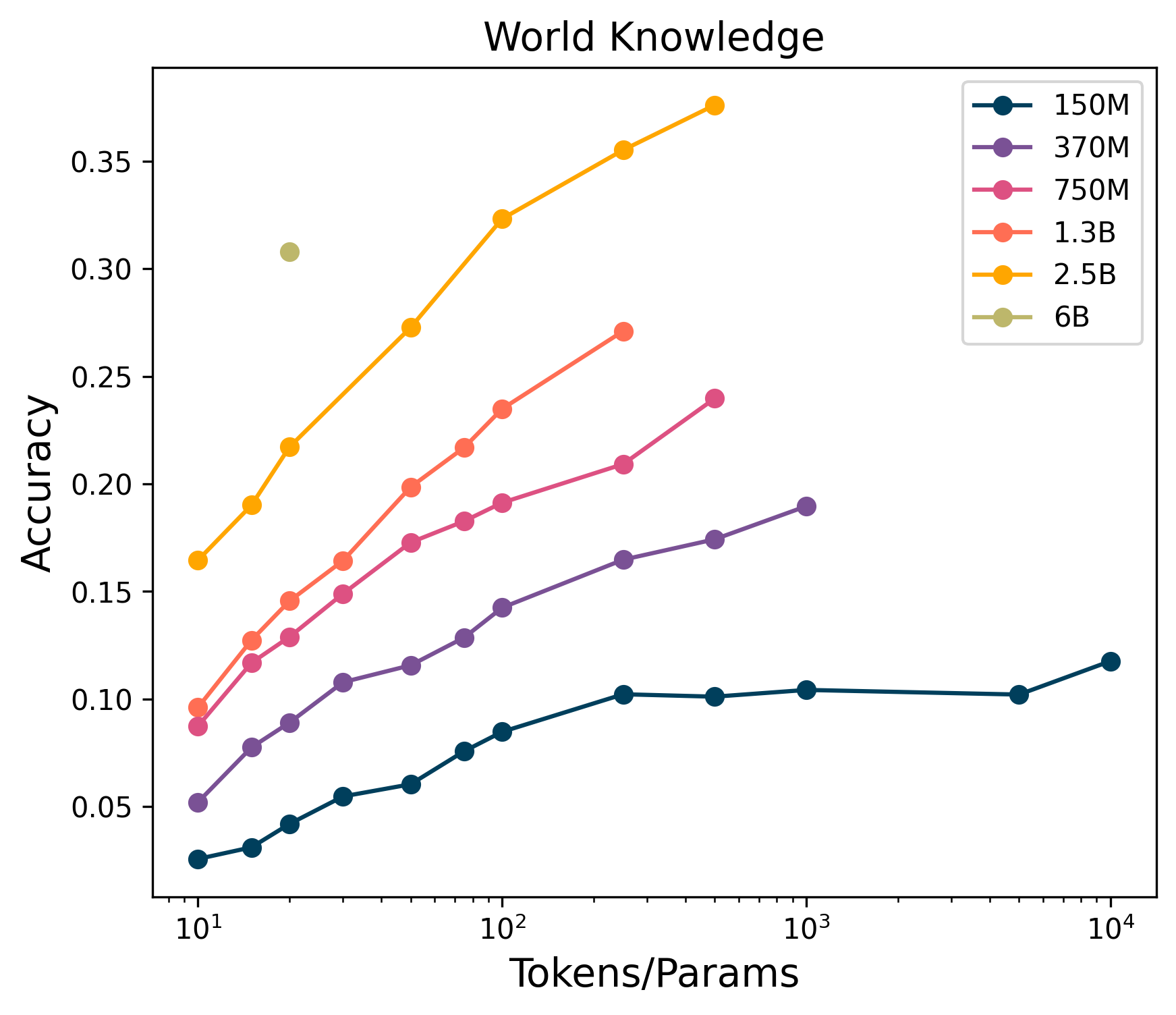}
                \caption{}
                \label{fig:worldknowledge}
        \end{subfigure}%
        \begin{subfigure}[b]{0.33\textwidth}
                \centering
                \includegraphics[width=1.00\linewidth]{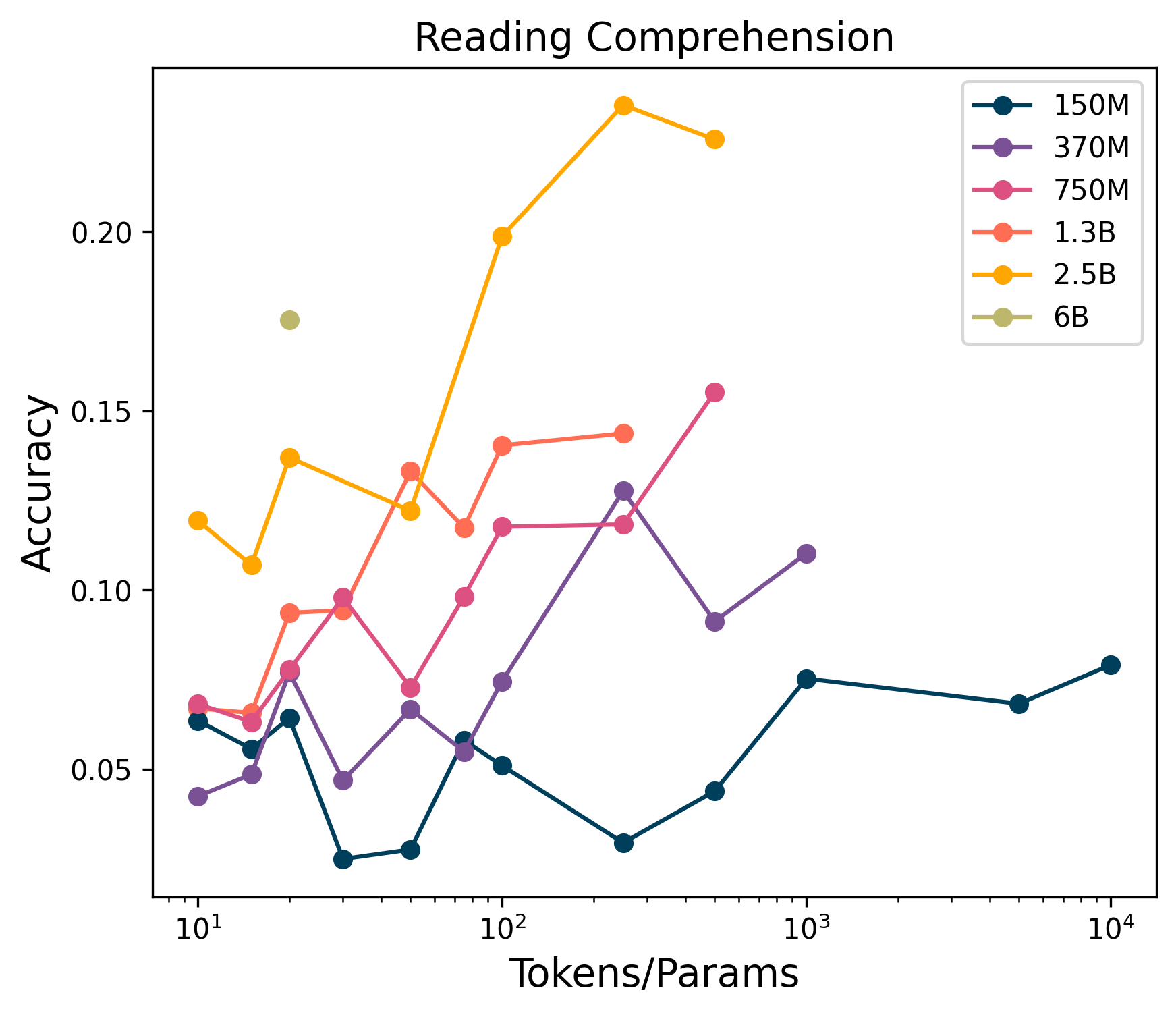}
                \caption{}
                \label{fig:readingcomprehension}
        \end{subfigure}%
        \begin{subfigure}[b]{0.33\textwidth}
                \centering
                \includegraphics[width=1.00\linewidth]{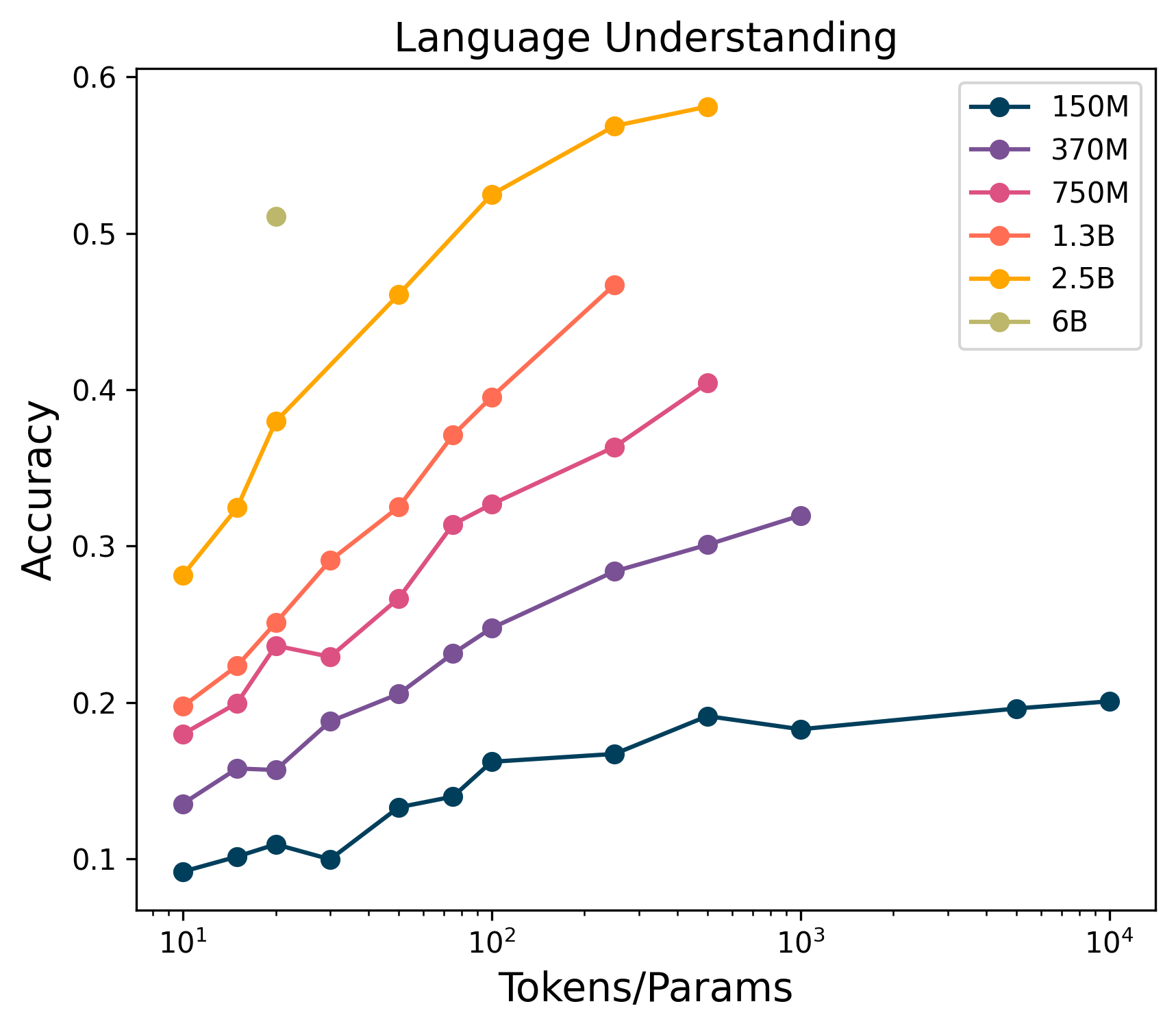}
                \caption{}
                \label{fig:languageunderstanding}
        \end{subfigure}%

        \begin{subfigure}[b]{0.33\textwidth}
                \centering
                \includegraphics[width=1.00\linewidth]{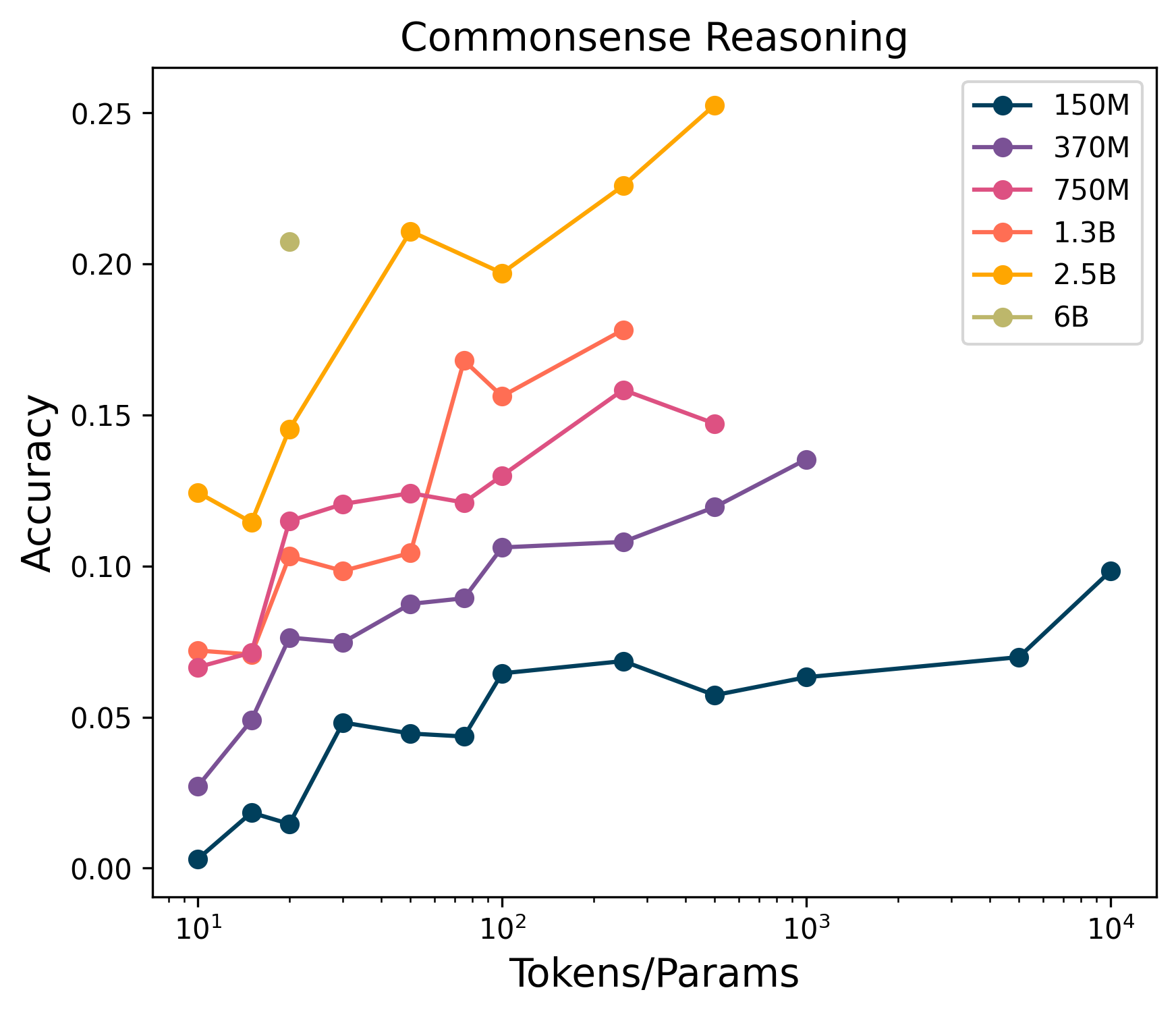}
                \caption{}
                \label{fig:commonsensereasoning}
        \end{subfigure}%
        \begin{subfigure}[b]{0.33\textwidth}
                \centering
                \includegraphics[width=1.00\linewidth]{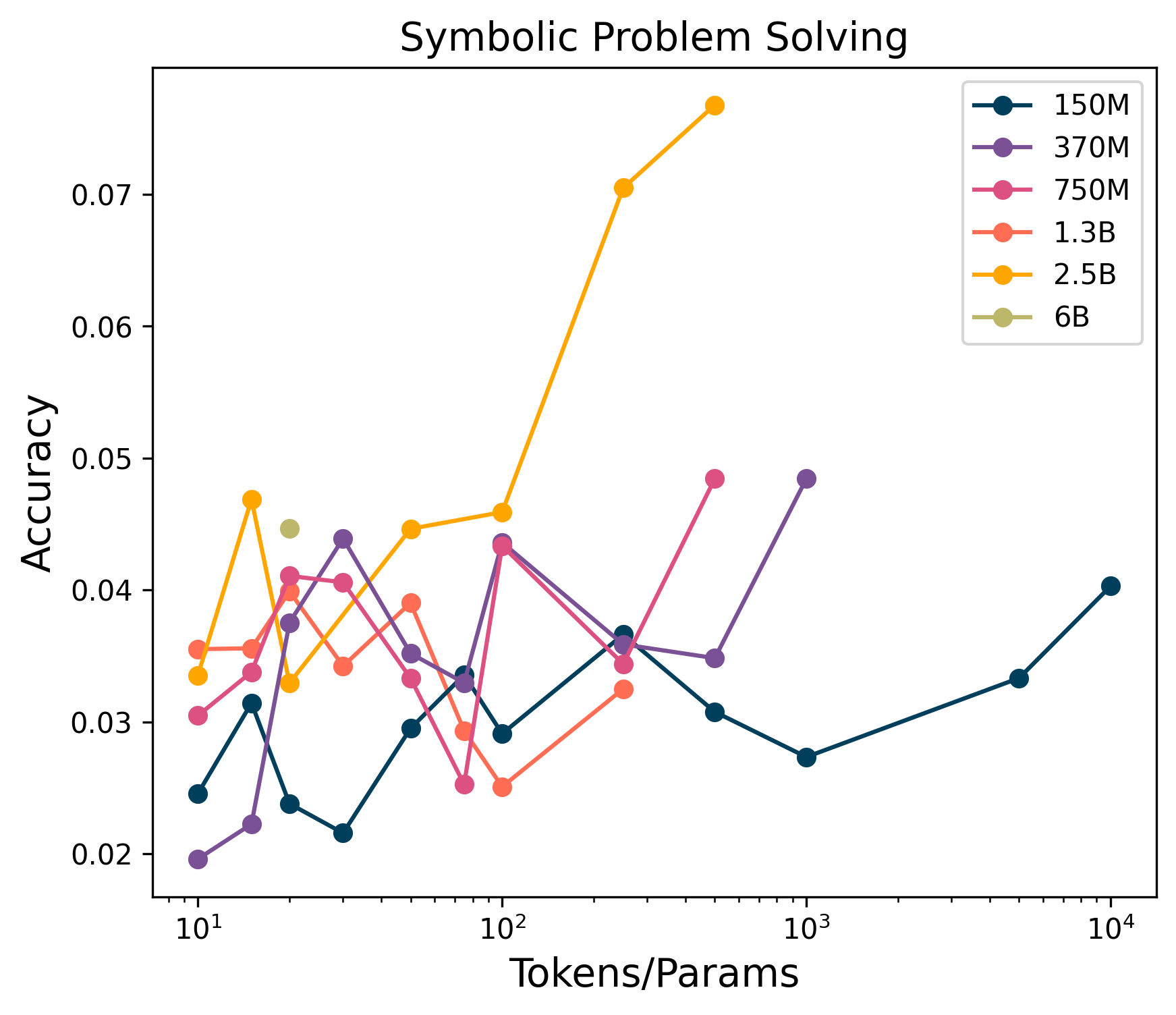}
                \caption{}
                \label{fig:symbolicproblemsolving}
        \end{subfigure}
        \caption{Per-category Gauntlet results.}
        \label{fig:gauntletcategories}
\end{figure}

We present more results from our experiments described in Section \ref{sec:experiments}. In Figure \ref{fig:gauntletcategories}, we show aggregate results for each Gauntlet category. Within each category, every task (the tasks are enumerated in Sec. \ref{sec:experiments}) is weighted equally and averaged after subtracting out baseline and normalizing so the maximum achievable accuracy is 1. Note that for some categories (e.g. Symbolic Problem Solving), all models achieve nearly zero performance on all tasks, resulting in low correlation between accuracy and token/parameter ratios (see \citet{Barton2024}).

\end{document}